\documentclass[sigconf]{acmart}

\renewcommand\footnotetextcopyrightpermission[1]{}
\usepackage{booktabs}
\usepackage{graphicx}
\usepackage{subcaption}
\usepackage{algorithmic}
\usepackage[ruled,linesnumbered,noend]{algorithm2e}
\usepackage{harpoon}
\usepackage{multicol}
\usepackage{multirow}
\usepackage{amsmath,amsfonts}
\usepackage{mathtools}
\usepackage{color}
\usepackage{enumitem}
\usepackage{footmisc}
\usepackage{verbatim}
\usepackage{balance}

\usepackage{comment}

\usepackage{listings}

\newcommand{\gm}[1]{\textcolor{red}{GM: #1}}

\newcommand{\rev}[1]{\textcolor{blue}{#1}}
\newcommand{\betterparagraph}[1]{\noindent\textbf{#1.}}
\newcommand{\proposed}{\textsc{SPION}}

\usepackage[ruled,linesnumbered,noend]{algorithm2e}
\SetKwComment{Comment}{$\triangleright$\ }{}
\SetCommentSty{mycommfont}
\SetKwFor{ParallelFor}{parallel for}{}{}
\SetKw{KwAnd}{and}
\SetKw{KwOr}{or}
\SetKw{KwBy}{by}
\usepackage{subcaption}
\AtBeginDocument{%
  \providecommand\BibTeX{{%
    \normalfont B\kern-0.5em{\scshape i\kern-0.25em b}\kern-0.8em\TeX}}}
    
\begin{document}

\title{\proposed{}: Layer-Wise Sparse Training of Transformer via Convolutional Flood Filling}


\author{Bokyeong Yoon}
\affiliation{%
  \institution{Sogang University}
  \streetaddress{35 Baekbeom-ro, Mapo-gu}
  \city{Seoul}
  \country{Republic of Korea}}
\email{bkyoon@sogang.ac.kr}

\author{Yoonsang Han}
\affiliation{%
  \institution{Sogang University}
  \streetaddress{35 Baekbeom-ro, Mapo-gu}
  \city{Seoul}
  \country{Republic of Korea}}
\email{han14931@sogang.ac.kr}

\author{Gordon Euhyun Moon}
\affiliation{%
  \institution{Sogang University}
  \streetaddress{35 Baekbeom-ro, Mapo-gu}
  \city{Seoul}
  \country{Republic of Korea}}
\email{ehmoon@sogang.ac.kr}

\renewcommand{\shortauthors}{Yoon and et al.}

\begin{abstract}
% \rev{irregular pattern and distribution in attention score matrix. layer-wise training}
Sparsifying the Transformer has garnered considerable interest, as training the Transformer is very computationally demanding. 
Prior efforts to sparsify the Transformer have either used a fixed pattern or data-driven approach to reduce the number of operations involving the computation of multi-head attention, which is the main bottleneck of the Transformer. 
% However, existing methods are still inevitably performance-limited in terms of ignoring important data points and increasing the model size.
% However, existing methods are still inevitably performance-limited in terms of failing to pay attention to important data points and increasing the model size.
% However, existing methods are inherently limited in terms of performance, as they encounter unavoidable challenges arising from the utilization of fixed patterns and the inclusion of additional parameters.
However, existing methods suffer from inevitable problems, such as the potential loss of essential sequence features due to the uniform fixed pattern applied across all layers, and an increase in the model size resulting from the use of additional parameters to learn sparsity patterns in attention operations.
In this paper, we propose a novel sparsification scheme for the Transformer that integrates convolution filters and the flood filling method to efficiently capture the layer-wise sparse pattern in attention operations. Our sparsification approach reduces the computational complexity and memory footprint of the Transformer during training.
% Our approach dynamically captures the pattern of sparsity in attention operations for each layer without the need to maintain additional parameters. 
% Our approach dynamically captures the sparsity pattern in attention operations for each layer, without maintaining additional parameters.
Efficient implementations of the layer-wise sparsified attention algorithm on GPUs are developed, demonstrating a new \proposed{} that achieves up to 3.08$\times$ speedup over existing state-of-the-art sparse Transformer models, with better evaluation quality.
% As efficiency of our new parallelization strategy is associated with the sequence length, our parallel GRU algorithm achieves significant performance improvement as the sequence length increases.
\end{abstract}

% \rev{Add "layer-wise MHA operation" as many as possible in Section 4 (the submitted paper does not include the term "layer-wise" in Section 4).}

% \rev{add MHA pattern and the number of non-zero elements are different and irregular in each layer. These patterns are kept changing during the training.  Also need to describe these problems in abstract.}

\maketitle
\pagestyle{plain}

\section{Introduction}
\label{sec:introduction}
% \rev{Add "layer-wise MHA operation" as many as possible in Section 4 (the submitted paper does not include the term "layer-wise" in Section 4).}

% \rev{add MHA pattern and the number of non-zero elements are different and irregular in each layer. These patterns are kept changing during the training.  Also need to describe these problems in abstract.}

% \rev{remove the term "long sequences"}

% Keywords that have to be included in introduction: compute-intensive, sprasification, optimization of MHA, reducing the number of operations associated with MHA, Convolutional Flood Fill, dynamically, flexible, layer-wise sparsification, limited memory space, resource-intensive with quadratic time and space complexity...

% \gm{\proposed{} is able to capture domain-specific sparsity pattern}

The Transformer is a state-of-the-art deep neural network developed for addressing sequence tasks, originally proposed by Vaswani et al. \cite{vaswani2017attention}.
One of the main advantages of the Transformer is that, given a sequence of input data points (e.g., a sentence of word tokens), it is able to compute the multi-head attention (MHA) operation in parallel, thereby quickly and accurately capturing long-term dependencies of data points.
% Therefore, the Transformer model is more powerful for longer input sequences compared to recurrent neural network-based models such as LSTM and GRU, as the training procedure of the latter is inherently sequential.
% Therefore, the Transformer model is known to be more efficient for longer input sequences compared to recurrent neural network-based models such as LSTM and GRU, as the latter inherently take the data points in the input sequence in sequential order.
% and have a vanishing gradient problem for long sequences.
However, as the sequence length increases, it is true that the overall computational cost and memory space required for training the Transformer also increase quadratically \cite{ainslie2020etc}.
Especially, a MHA sub-layer in the Transformer occupies a substantial portion of the total execution time and becomes the main bottleneck as the sequence length increases.
% \begin{figure}[!ht]
% \centering
% \includegraphics[width=0.49\textwidth]{figures/breakdown.pdf}
% \caption{Breakdown of elapsed time and operation counts for processing the forward propagation of an encoder layer in Transformer based on the sequence length \gm{What about remove or move this figure to Section 2.}}
% \label{fig:breakdown}
% \end{figure}
% \index{figure}
% In addition, 
The MHA operation requires a large number of dot-product operations to compute the similarity between all data points in the sequence.
However, the dot-product operation inherently has limitations in memory bandwidth since it performs only two floating-point operations for each pair of data elements read from memory. Therefore, reducing the number of operations involving dot-product operations is an important consideration for accelerating the training of the Transformer.
In addition, the attention matrices used in the MHA computation have two main challenges.
% In practice, the attention matrices contain non-zero entries with an irregular distribution which leads to a load imbalance and a reduced level of exploitable parallelism.
First, as the pattern of non-zero entries in the attention matrices varies across different layers, the irregular access of these non-zero entries within the matrices leads to an inefficient utilization of global memory bandwidth and low cache hit rates while performing dot-product operations.
Second, the irregular distribution of non-zero entries in attention matrices across different layers results in load imbalance and a reduced level of exploitable parallelism.
% In order to reduce the computational complexity of the Transformer, 
Hence, in order to mitigate computational complexity and improve data locality of the Transformer, several approaches have addressed the sparsification of computations associated with the MHA operation \cite{zaheer2020big,beltagy2020longformer,roy2021efficient,zhang2021poolingformerLD,wang2020linformer}.
However, previous approaches suffer from two primary limitations.
% that stem from the use of fixed patterns and the requirement for additional parameters.
Firstly, when the Transformer adopts identical fixed patterns of non-zero entries in the attention matrices across all layers during training, it becomes difficult to effectively capture key features within the sequence.
% Secondly, when the Transformer employs additional parameters to learn the sparsity pattern in the attention score matrices, the size of the model increases, resulting in computational overhead.
Secondly, when the Transformer employs additional parameters to learn the sparsity pattern in the attention matrices, both the model size and the computational overhead increase.
% To address the limitations of previous approaches, our key contribution is to propose a domain-specific efficient sparse attention algorithm that efficiently captures sparsity patterns in MHA during training and significantly reduces memory consumption without loss of accuracy.
% To address the limitations of previous approaches, we focus on a specialized sparse attention algorithm that efficiently captures sparse patterns in MHA during training and significantly reduces memory consumption without loss of accuracy.
To address the limitations of previous approaches, we focus on a specialized sparse attention algorithm that not only significantly reduces memory consumption but also efficiently captures sparse patterns across various types of Transformers and datasets.
% and significantly reduces memory consumption without loss of accuracy.

% required to learn additional parameters.

% However, previous approaches suffer from inevitable problems, such as the loss of important features in the sequence using a fixed pattern, an increase in the model size,

% In this paper, we present a new sparsity-aware Transformer (called \textbf{\proposed{}}) that dynamically reduces the time and the number of operations required to train very long sequences without additional memory consumption.
% In this paper, we apply layer-wise attention head pruning on All-attention Transformer so that the entire computation and the number of ...
In this paper, we present a new sparsity-aware layer-wise Transformer (called \textbf{\proposed{}}) that dynamically captures the sparse pattern in the MHA operation for each layer.
% Our approach reduces both the time and the number of operations required to train long sequences without additional parameters.
% achieves efficient sparsification of the MHA operation without any loss in model quality.
\proposed{} judiciously explores the sparsity pattern in the MHA operation based on a novel convolutional flood filling method.
% layer-wise MHA computation
% sparsifies the MHA operation without any loss in model quality.
% Our sparsity pattern generator considers the shape of sparsity pattern, the degree of sparsity, and the spatial locality within the attention matrix.
% To precisely identify the pattern of sparsity in the attention matrix, 
To precisely detect the characteristics of the sparse pattern in the attention matrices, \proposed{} captures the shape of sparse pattern by utilizing a convolution filter and the degree of sparsity through a flood filling-based scheme. 
% While performing the sparse MHA, our \proposed{} also takes into account the data locality by forming a blocked sparsity pattern matrix.
During the generation of the sparse pattern for each layer, our \proposed{} constructs a sparsity pattern matrix with a blocked structure, enhancing data locality for the MHA operation that involves sparse matrix multiplication.
% also takes into account the data locality by constructing a sparsity pattern matrix with a blocked structure.
% performing the sparse MHA, our \proposed{} also takes into account the data locality by forming a blocked sparsity pattern matrix.
% is also considered by forming a block sparse matrix.
Furthermore, by capturing layer-specific sparse pattern for each layer, \proposed{} performs layer-wise MHA computations iteratively until convergence is achieved.
Our sparsification technique reduces both the time and the number of operations required to train long sequences without using additional parameters.
% Furthermore, we develop an efficient GPU implementation of the sparse MHA to achieve high performance with quality of results.
% Furthermore, as the sparse MHA operations account for a large portion of training \proposed{}, we develop an efficient GPU implementation of the sparse MHA to achieve high performance with quality of results.
Moreover, since the sparse MHA operations contribute significantly to the overall training workload, we develop an efficient GPU implementation for sparse MHA to achieve high performance with quality of results.
% Moreover, since the sparse MHA operations significantly account for training workload, we develop an efficient GPU implementation for sparse MHA to achieve high performance with quality of results.

% Our approach reduces both the time and the number of operations required to train long sequences without additional parameters.
% achieves efficient sparsification of the MHA operation without any loss in model quality.

We conduct an extensive comparative evaluation of \proposed{} across various classification tasks, including image and text classification, as well as document retrieval involving lengthy sequences.
Experimental results demonstrate that our \proposed{} obtains up to 10$\times$ lower operations 
% \rev{(involving MHA computation?)} 
% \bk{10x not in experiments case. larger in experiments case. it is about 20 $\times$} 
% \gm{what do you mean by real case?} 
and up to 3.08$\times$ training speedup compared to existing state-of-the-art sparse Transformers.

The paper is organized as follows. Section \ref{sec:background} presents the background on Transformer, sparse attention, flood fill algorithm, and related prior work. Section \ref{sec:motivation} describes the motivation behind our \proposed{}. Section \ref{sec:model} presents an overview and details of our \proposed{} implementation. Section \ref{sec:experiment} extensively compares the performance of \proposed{} with various state-of-the-art sparse Transformer models.

% These observations suggest that the MHA operation significantly affects the model's overall operations, and reducing the number of MHA operations can effectively decrease the model's overall training time.

% The paper is organized as follows. In the next section, we present background on Transformer, sparse MHA, flood fill agorithm and related prior work. In Section 3, we present a motivation of our algorithm (high-level overview of the ALO-NMF algorithm). Section 4 provides details of ALO-NMF for multi-core CPUs and GPUs. In Section 5, we compare the data movement cost for ALO-NMF and the orig- inal FAST-HALS algorithm. Thereafter, we discuss the approach to tile size selection based on data movement analysis. Section 6 com- pares the performance of ALO-NMF with existing state-of-the-art parallel implementations.
\section{Background and Related Work}
\label{sec:background}

% In this section, we begin by providing the background on Transformer and sparse attention mechanism. 

% To begin, we will first provide an overview of the Transformer architecture and its key component, the attention mechanism. Following that, we will introduce the concept of sparse attention and explore the efficient Transformer models that have incorporated it.

%We then analyze the sparsity of results after the attention operation and identify common patterns that emerge. 
%This analysis is crucial in understanding the computational efficiency of the Transformer model and developing effective optimization techniques.

\subsection{Transformer}
% \betterparagraph{Transformer}

% This paper aims to address the issue of computational optimization in the attention mechanism of the Transformer model. 
% The traditional Transformer consists of an encoder and a decoder, which are composed of multiple multi-head attention (MHA) layers and feed-forward layers.
% The Transformer consists of multiple multi-head attention (MHA) layers and feed-forward layers.
% Unlike Recurrent Neural Network-based models that sequentially take each data point in a sequence as input, the Transformer model takes the entire sequence of data points as input.
% The entire input sequence is then used to perform the attention mechanism, computing the similarity between each data point and all other data points in the sequence.

%Therefore, the attention layers are crucial component in the learning process of the Transformer model.

% One of the Transformer variants, the encoder-only Transformer, is widely used for various classification tasks using text and image datasets \cite{tay2022efficient,devlin2018bert,dosovitskiy2020image}.
The encoder-only Transformer, which is one of the variants of the Transformer model, is widely used for various classification tasks using text and image datasets \cite{tay2022efficient,devlin2018bert,dosovitskiy2020image}.
% One of the Transformer variants, the encoder-based Transformer, is widely used to solve various classification tasks using text and image datasets. \cite{devlin2018bert,dosovitskiy2020image}.
% Algorithm \ref{alg:encoder} shows pseudo-code for the Transformer encoder, which is composed of multiple multi-head attention (MHA) layers (lines 7-9) and feed-forward layers (lines 12-14).
% Algorithm \ref{alg:encoder} shows pseudo-code for the encoder layers of Transformer, where each encoder layer consists of a multi-head attention (MHA) layer (lines 7-9) and a feed-forward layer (lines 12-14).
Algorithm \ref{alg:encoder} shows pseudo-code for the forward propagation of the original encoder-only Transformer, where each encoder layer consists of a MHA sub-layer (lines 2-10) and a feed-forward sub-layer (lines 11-12).
\begin{algorithm}[!ht]
\caption{Forward propagation of encoder layer in encoder-only Transformer}
\label{alg:encoder}
% \footnotesize
% \scriptsize
\SetAlgoLined
\DontPrintSemicolon
\KwIn{$E \in \mathbb{R}^{L \times D}$: a set of embedding vectors for an input sequence, $L$: length of input sequence , $D$: embedding size for each data point, $H$: the number of heads, $W^{Q}$, $W^{K}$, $W^{V}$, $W^{O}$, $W^{F}$, $W^{E}$: weight parameters, $N$: the number of encoder layers}
% $D$: embedding size for each data point in the input sequence
% \KwOut{$E$: ($L\times D$) embedding vectors produced by the last encoder layer}

% $token\_output \gets token\_embedding(X)$

% $position\_output \gets position\_embedding(X)$

% $encoder\_input \gets token\_output + position\_output$
% $X_{embed} \gets token\_embed(X) + pos\_embed(X)$
\For{$encoder\_layer = 0$ \KwTo $N-1$}{
\Comment*[r]{MHA sub-layer}
$X \gets LayerNorm(E)$
% $X \gets LayerNorm(E)$ \Comment*[r]{\textcolor{blue}{MHA sub-layer}}

% $Q, K, V \gets Z\times W_Q, Z\times W_K, Z\times W_V$
$Q \gets X\times W^{Q}, K \gets X\times W^{K}, V \gets X\times W^{V}$

$Q_{0,...,H-1}$, $K_{0,...,H-1}$, $V_{0,...,H-1}$ $\gets split(Q, K, V)$
% \text{Splitting each $Q$, $K$, and $V$ into $H$ attention heads}$

% \Comment*[r]{XXX}
\ParallelFor{$i = 0$ \KwTo $H-1$}{

    % $R_i \gets Q_i \times K_{i}^T$
    $A_{i}^{r} \gets Q_i \times K_{i}^T$

    $A_{i}^{s} \gets softmax(A_{i}^{r} / \sqrt{D/H})$

    $A_{i}^{c} \gets A_{i}^{s} \times V_i$

}

$A \gets concatenate (A^{c}_{0}, ... , A^{c}_{H-1})$

% \Comment*[r]{\textcolor{blue}{Feed-forward sub-layer}}
$O \gets dropout(A \times W^{O}) + E$
% $O \gets dropout(A \times W^{O}) + E$ \Comment*[r]{\textcolor{blue}{Feed-forward sub-layer}}

\Comment*[r]{Feed-forward sub-layer}
% $A \gets dropout(A) + X$
% $M \gets LayerNorm(O)$
$F \gets ReLU(LayerNorm(O) \times W^{F})$

$E \gets dropout(F \times W^{E}) + O$

% $attn\_mat \gets Q \times K^T$

% $attn\_score \gets softmax(attn\_mat/\sqrt{D/H})$
% $S \gets softmax((Q \times K^T)/\sqrt{D/H})$

% $attn\_output \gets attn\_score \times V$
% $attn\_output \gets S \times V$

% $attn\_output \gets dropout(attn\_output) + X$

% $ln2\_output \gets LayerNorm(attn\_output)$

% $ff1\_output \gets ln2\_output \times W_{1}$

% $ff1\_activation \gets ReLU(ff1\_output)$

% $ff2\_output \gets ff1\_activation \times W_{2}$

% $enc\_input \gets dropout(ff2\_output) + attn\_output$
}

% $output \gets LayerNorm(E)$

\end{algorithm}
For each encoder layer, the input embedding $E$ is transformed into $X$ by applying layer normalization (line 2), and the query ($Q$), key ($K$), and value ($V$) are obtained by performing linear transformations on the embedding $X$ (line 3).
Hereafter, we denote $L$, $D$ and $H$ as the length of input sequence, the size of the embedding for each data point in the sequence, and the number of heads, respectively. To efficiently perform MHA computation, each $Q \in \mathbb{R}^{L \times D}$, $K \in \mathbb{R}^{L \times D}$, and $V \in \mathbb{R}^{L \times D}$ matrix is divided into $H$ sub-matrices (multi-heads) along the $D$ dimension (line 4).
% Hence, the size of $Q$, $K$, and $V$ is $H\times L \times (D/H)$.
The utilization of multi-heads indicates that the embedding of the input sequence is divided into multiple subspaces. This division enables the model to perform MHA in parallel (lines 5-8) while precisely capturing distinct key features in each subspace.
% The use of multi-heads allows for the division of the vector representing the input sequence into multiple subspaces, facilitating the reference of different information at the same time, and ultimately leading to a more multifaceted representation.
% In line 4, weights $W_q$, $W_k$, and $W_v$ are used to generate $Q$, $K$, and $V$, respectively, through a multiplication operation with layer-normalized input $E$.
% The resulting vectors $Q$ and $K$ are then used to compute the matrix multiplication kernel (GEMM), which captures the similarity between the two input sequence vectors. 
% The resulting value is then scaled by $1/\sqrt{D/H}$ to prevent the gradients of the softmax function from approaching zero, and the softmax function is applied row-wise to obtain the attention score($attn\_score$). 
% Finally, the GEMM operation between the $attn\_score$ and the vector Value$(V)$ is performed to obtain an embedding representation for each input sequence vector.
% The computation within $i$ loops involves matrix-matrix multiplication (GEMM) to calculate the similarity between query $Q_i \in \mathbb{R}^{L \times (D/H)}$ and key $K_i \in \mathbb{R}^{L \times (D/H)}$ for head $i$ (line 5).
The loop in lines 5-8 performs the MHA computation as formulated in Equation \ref{eq:attention}.
% \begin{equation}
% \label{eq:attn}
% \begin{aligned}
% &A^{c} = softmax((Q \boldsymbol{\cdot} K^T)\sqrt{(D/H)})\boldsymbol{\cdot} V\\
% \end{aligned}
% \end{equation}
\begin{equation}
\label{eq:attention}
% \begin{aligned}
A = softmax\left(\frac{Q \times K^T}{\sqrt{(D/H)}}\right) \times V
% \end{aligned}
\end{equation}
The computation within $i$ loops involves matrix-matrix multiplication (GEMM) to obtain the raw attention score matrix $A_{i}^{r}$, which captures similarity between query $Q_i \in \mathbb{R}^{L \times (D/H)}$ and key $K_i \in \mathbb{R}^{L \times (D/H)}$ for head $i$ (line 6). 
The softmax function is then applied to the matrix $A_{i}^{r}$, which is scaled by $1/\sqrt{D/H}$ (line 7). This scaling is performed to prevent the gradients of the attention score matrix $A_{i}^{s}$ from approaching zero.
Thereafter, another GEMM operation is performed between matrices $A_{i}^{s}$ and $V_{i} \in \mathbb{R}^{L \times (D/H)}$ to obtain the complete attention matrix $A_{i}^{c}$ for head $i$ (line 8).
% After computing the attention for each head, all matrices $A_{0}^{c}$ through $A_{H-1}^{c}$ are concatenated to form the final attention matrix $A$ (line 9). 
After computing the attention for each head, all matrices $A_{0}^{c}$ through $A_{H-1}^{c}$ are concatenated to form the final attention matrix $A$, and the input embedding $E$ used at the current encoder layer is added to the attention matrix (lines 9 and 10).
Then, the attention matrix is passed through the feed-forward sub-layer to produce new embedding vectors $E$, which is then fed into the next encoder layer (lines 11 and 12).
% From the number of operations standpoint, the MHA computation (lines 5-8) requires XYZ 
% For the computation of MHA (lines 5-8), the number of operations required is is XYZ, indicating that the number of operations increases exponentially as the length of input sequence increases.
% In order to decrease the computational complexity required for processing each encoder layer, it is evident that 
In terms of computational complexity, it is clear that the main bottleneck of the encoder layer is associated with processing the MHA sub-layer (lines 2-10).
% More specifically, the number of operations required for the loop in line 5 is \gm{XYZ}, indicating an exponential increase in operations as the length of the input sequence ($L$) grows.
%$L^{2}(2D-1)+L(3L-1)+LD(2L-1)$
More specifically, the number of operations required for computing the attention matrix $A$ for head $i$ (loop in line 5) is $2L^2(2D+1)-L(D+1)$, indicating a quadratic increase in operations as the input sequence length ($L$) increases.

% For the computation of MHA (lines 5-8), the number of operations required is \gm{XYZ}, indicating an exponential increase in operations as the length of the input sequence grows.

% \gm{@BK: I will come back here later. the number of operations for running entire encoder layer vs. MHA, computational complexity (as $L$ increase, size of $A_{i}^{r}$ and $A_{i}^{s}$ increases..)} 

% From a computational complexity standpoint, the MHA computation has quadratic complexity as the computational complexity for line 6, line 7, and line 8 is $O(L^2D)$, $O(L^2)$, and $O(L^2D)$, respectively.

% The computational complexity for line 6, line 7, and line 8 is $O(L^2D)$, $O(L^2)$, and $O(L^2D)$, respectively.
% From a computational complexity standpoint, the complexity of MHA exponentially increases as the sequence length $L$ increases.

% Specifically, the computational complexity of calculating the raw attention score $A^r$ is $O(L^2D)$, while the computational complexity of multiplying $A^s$ and $V$ is also $O(L^2D)$. 
% And computational complexity of softmax function is $O(L^2)$.
% Therefore, the computational complexity of the attention mechanism increases proportionally to the square of the input sequence length $L$.
%Consequently, the attention mechanism's computational complexity grows quadratically relative to the input sequence length, $L$.

\betterparagraph{Sparsification of MHA}
%To mitigate quadratically increasing computational complexity by the sequence length  there has proposed efficient transformer that uses sparse attention. sparse attention is based on the concept that not all data points in input sequence has to be computed. If we perform similarity calculations to only those data points that we consider relevant, and this does not affect the model's accuracy, we can achieve efficient reductions in both computational and memory complexity. Sparse Attention refers to the approach of performing attention operations only on the relevant data points in the input sequence. 
% Efficient Transformers that utilize sparse attention have been proposed to address the limitation of quadratically increasing computational complexity as the sequence length grows\cite{child2019generating,beltagy2020longformer,zaheer2020big}.
% The concept of sparse attention is based on the idea that not all data points in the input sequence need to be computed. 
% Several efficient sparse attention techniques have recently been proposed to reduce the number of operations required for computing the raw attention matrix $A^{r}$, given the long input sequences.
Given the long input sequences, several sparse attention techniques have been proposed to reduce the computational complexity involved in computing the raw attention matrix $A^{r}$.
% The basic intuition behind sparse attention techniques is that the long input sequence can be effectively represented by a subset of its data points.
The basic intuition behind sparse attention techniques is that a subset of data points in the long sequence can effectively represent the entire input sequence.
% In other words, to reduce the number of operations for computing the raw attention matrix $A^{r}$, only the strongly correlated elements (i.e., data points) in the $Q$ and $K$ can be used.
% In other words, to reduce the computational workload involved in computing the raw attention matrix $A^{r}$, it is possible to utilize solely the highly correlated elements (i.e., data points) in the $Q$ and $K$.
In other words, to reduce the computational workload of computing the raw attention matrix $A^{r}$, only the highly correlated necessary elements (i.e., data points) in $Q$ and $K$ can be utilized.
% By exclusively performing similarity calculations on the relevant data points, while maintaining model accuracy, computational and memory complexity can be reduced.
% Sparse attention operates solely on the relevant data points in the input sequence.
% \begin{equation}
% \label{eq:spattn}
% \begin{aligned}
% &SpAttnOutput = SparseSoftmax((Q \boldsymbol{\cdot} K^T) \odot SpPttn/\sqrt{(D/H)})\boldsymbol{\cdot} V\\
% \end{aligned}
% \end{equation}
% With the Sparse Attention, computational complexity and memory usage can be significantly reduced.
Therefore, when employing sparse attention, the number of operations required to compute $A_{i}^{r}$ for each head $i$ is $C\times(2D-1)$, where $C$ is the number of critical data points in the long sequence, whereas the number of operations required for the original dense $A_{i}^{r}$ is $L^{2}\times(2D-1)$. 

\subsection{Flood Fill Algorithm}
\label{subsec:flood_fill}
% \gm{@BK: write the background of flood fill algorithm using 1-2 paragraphs. Use the terms like "dynamic programming", "dynamic", "dynamically" as much as possible.}
The flood fill scheme is originally developed to determine the area connected to a given cell/pixel in a multi-dimensional array and a bitmap image \cite{goldman1990graphics}. Starting from a given seed element, 
% the algorithm dynamically explores all the neighboring points surrounding the current element.
% For each step, 
the algorithm determines whether to continue or stop the fill operation based on conditional statements that evaluate the properties of the current element and its surrounding element.
% \gm{@BK: Specify and describe the condition that makes continuing the fill operation and the other condition that makes stopping the fill operation. For example, "\textit{If ... is satisfied, the fill operation is performed; otherwise, it is avoided.}"} 
If the current element is not at the end of the data structure and the neighboring elements have not been explored yet, the algorithm continues; otherwise, it is avoided.
% \textcolor{blue}{To ensure the dynamic exploration of all possible paths in the entire array, this algorithm recursively applies the fill operation to the current point if the neighboring points satisfy the condition.}
Furthermore, the flood fill algorithm dynamically controls the fill operation with a constraint on small values of the elements. If the value of current element is very small, it prevents the fill operation.

\subsection{Related Work on Sparse Attention}
\label{subsec:related_work}

% Many previous efforts at parallelizing LDA have sought to relax CGS and use approximations because strict CGS im- poses constraints on efficient parallelization. 

Many previous efforts to achieve efficient Transformers have sought to sparsify the MHA operation both before and during model training/fine-tuning.
\betterparagraph{Fixed Sparse Pattern}
One of the strategies for performing sparse MHA is to use a predetermined sparsity patterns, where only specific data points in the input sequence are selected before training the model. In this approach, only these selected data points are exclusively used to perform the MHA operation.
% Sparse attention mechanism are then exclusively performed on these selected data points. 
% Sparse Transformer\cite{child2019generating}
% Several models, such as Longformer\cite{beltagy2020longformer}, ETC\cite{ainslie2020etc}, and BigBird\cite{zaheer2020big}, adapt the sliding windows approach in which the attention operations are solely performed on the neighboring elements (data points) to the left and right of the diagonal elements in the attention matrix.
% such as Sparse Transformer\cite{child2019generating}, Longformer\cite{beltagy2020longformer}, ETC\cite{ainslie2020etc}, and BigBird\cite{zaheer2020big}
Several variants of the encoder-only Transformer model adapt the sliding windows approach in which the attention operations are performed using only the neighboring data points (rows) in the matrices $Q$ and $K$.
The Sparse Transformer \cite{child2019generating} originally employs the sliding windows attention to sparsify the MHA operation.
% In the Sparse Transformer, sparsity is achieved by calculating the similarity only between the neighbors of the data points in the $Q$ and $K$.
% The Longformer, as a derivative of Sparse Transformer, 
% The Longformer is an extension to the Sparse Transformer \cite{child2019generating} that introduces the dilated sliding windows, which extends the receptive field for similarity calculation while preserving the same computational complexity.
The Longformer \cite{beltagy2020longformer} is an extension to the Sparse Transformer and introduces dilated sliding windows, which extend the receptive field for computing similarity by skipping one data point at a time while performing the sliding windows attention.
% while preserving the same computational complexity.
Furthermore, ETC \cite{ainslie2020etc} and BigBird \cite{zaheer2020big} incorporate global attention that performs similarity calculations between a given data point and all other data points in the input sequence.
ETC introduces three types of fixed sparsity patterns for attention, called global-global, global-local, and local-global.
BigBird further proposes random attention, which randomly selects data points for similarity computation.

% \rev{Most recently, the LSG Attention \cite{condevaux2023lsg} extends ... for fine-tuning the Transformer model and outperforms ... on text classification task.}
% \bk{Most recently, the LSG Attention \cite{condevaux2023lsg} extends pretrained RoBERTa\cite{liu2019roberta} for a text classification task, fine-tuning the LSG Attention model and outperforming Longformer \cite{beltagy2020longformer} and BigBird \cite{zaheer2020big}.}

% In addition, BlockBert \cite{qiu2019blockwise} performs the attention operation on a single block for each block-wise row and conducts permutation to vary the position of the block for each block-wise row in each head.
% \gm{By utilizing these non-learnable fixed sparsity patterns of MHA computation, one key advantage is the reduction in computational overhead. Once the pattern is predefined, there is no need to learn patterns during the training process. This saves computational resources and eliminates the additional complexity associated with dynamically determining the sparse connections.}
% The main advantage of utilizing these fixed sparsity patterns is the reduction in computational overhead.
The main advantage of sparsifying the MHA operation before training is the reduction in computational overhead.
% As the fixed sparsity patterns are applied to all MHA computations in the entire training procedure, this approach enables reducing memory footprint and eliminating the additional computational complexity required to dynamically determine the sparsity patterns during training.
As the fixed sparsity patterns are applied to all MHA computations in the entire training procedure, this approach enables reducing the overall memory footprint compared to the original dense MHA operation.
% required to train the model.
However, the primary problem with a fixed pattern is that it is not sufficient to fully capture the dependencies present in the input sequence.
For example, depending on the types of dataset and task, the critical data points in the long input sequence can vary. Therefore, predefined static sparse patterns may lose the fine-grained important features and dependencies in the sequence during training. 
% \gm{(need to describe the details of what features/information can be lose)}
% \bk{Therefore, if the fine-grained important features and dependencies are not included in the predefined static sparse patterns, the model will not operate for that elements, losing the information of that element could have during training.}
%\cite{vyas2020fast,wang2020cluster}
% \betterparagraph{Data-driven Sparsity Patterns}
% \betterparagraph{Sparsifying during Training}

\betterparagraph{Data-driven Sparse Pattern}
% (including sparsify-during-fine-tuning)
% Sparsity patterns in the MHA operation can also be captured through learning \gm{learning what?}. 
% By leveraging data-driven approaches, functions \gm{which functions?} can be learned to cluster and sort data points within the input sequence, leading to the generation of sparse patterns in MHA operation. 
Sparse patterns in the MHA operation can also be generated by leveraging a data-driven approach, which clusters and sorts the data points of the input sequence during training.
% to generate the sparsity patterns in MHA during training.
% learns the sparsity patterns in attention 
% , functions \gm{which functions?} can be learned to cluster and sort data points within the input sequence, leading to the generation of sparse patterns in MHA operation. 
% This data-driven approach allows for adaptability and flexibility in capturing relevant dependencies while reducing computational complexity. 
% Reformer\cite{kitaev2020reformer}, for instance, utilizes locality-sensitive hashing (LSH) to calculate hash-based similarities and clusters each data point into chunks. 
For example, Reformer \cite{kitaev2020reformer} utilizes locality-sensitive hashing to calculate a hash-based similarity and cluster multiple data points into chunks. 
Similarly, Routing Transformer \cite{roy2021efficient} performs k-means clustering given data points.
In the process of training the model, these clustering-based attention approaches also require learning additional functions to effectively identify the relevant dependencies of data points in the input sequence.
% \gm{Include more related work on sparsifying during fine-tuning in this paragraph}
Wang et al. \cite{Wang2019StructuredPO} reduce the number of operations required for MHA computation by parameterizing weight matrices involved in MHA using low-rank matrix factorization and eliminating rank-1 components during model training.
% During the process of fine-tuning, Peer et al. \cite{Peer2021GreedylayerPS} and Lagunas et al. \cite{Lagunas2021BlockPF} employ the computed values obtained from the parameters of the pre-trained model to reduce the model size by pruning the insignificant layers and blocks, respectively.
Peer et al. \cite{Peer2021GreedylayerPS} and Lagunas et al. \cite{Lagunas2021BlockPF} reduce the size of the fine-tuning model by removing the insignificant layers and sub-blocks in attention matrix based on the parameters from the pretrained model.
However, even though utilizing data-driven sparsification techniques during training produces a better quality model, this approach requires additional parameters and operations to learn the sparse patterns, resulting in larger memory space and higher computational cost. 
\section{Motivation: Analysis of Sparse Patterns in MHA}
\label{sec:motivation}

\begin{figure}[!ht]
\centering
\includegraphics[width=0.49\textwidth]{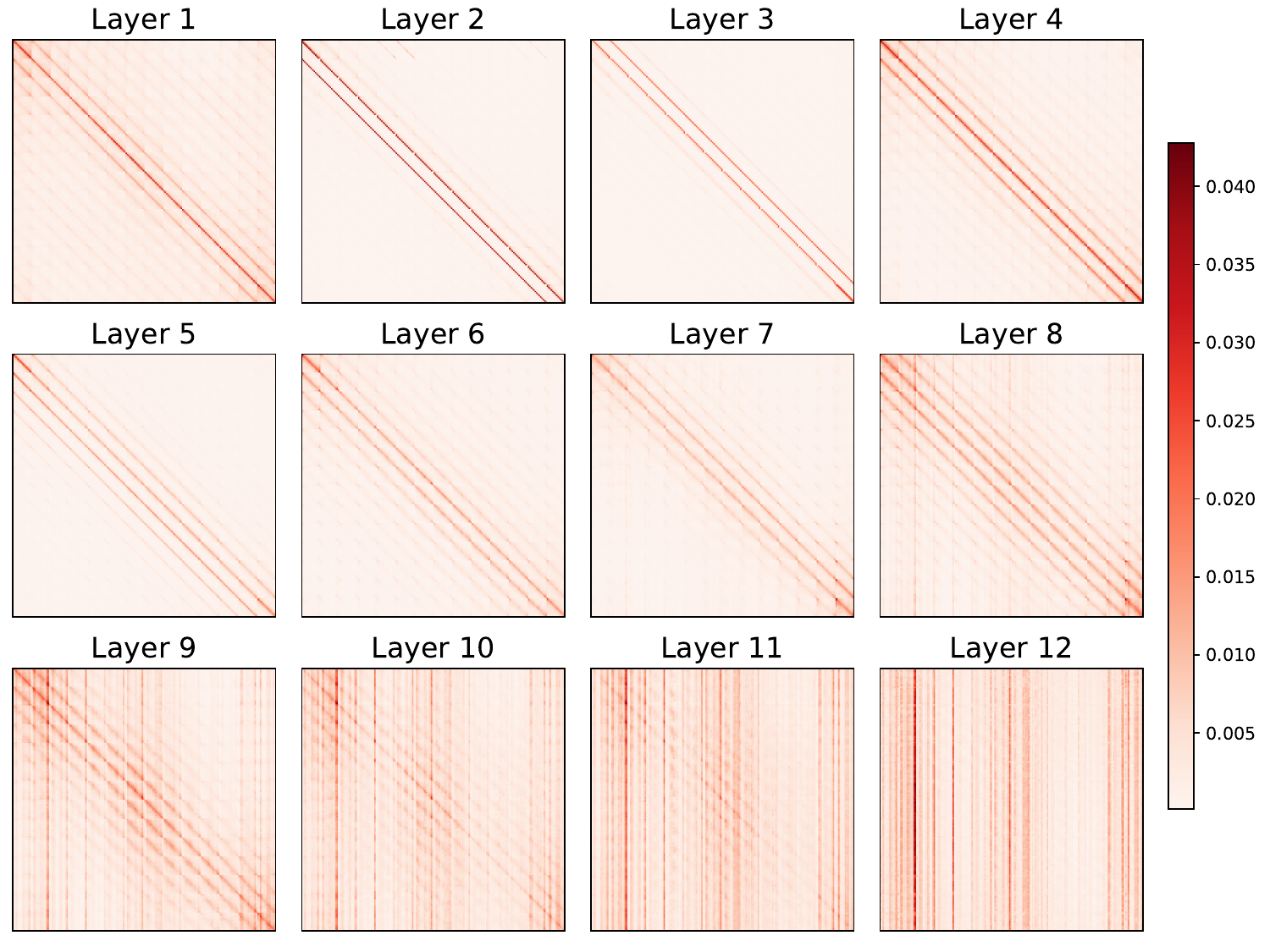}
\caption{Sparsity patterns in the attention score matrices $A^s$ across different encoder layers during training of the encoder-only Transformer for image classification. 
% \rev{Reduce the number of sub-figures?}
}
\label{fig:motivation}
\end{figure}
\index{figure}

% \begin{figure}[!ht]
% \centering
% \includegraphics[width=0.49\textwidth]{figures/motivation.pdf}
% \caption{Sparsity patterns in the attention score matrices across different encoder layers during fine-tuning of the encoder-only Transformer for image classification. \rev{Reduce the number of sub-figures?}}
% \label{fig:motivation}
% \end{figure}
% \index{figure}

% \begin{figure*}
% \centering
% \begin{subfigure}{0.95\textwidth}
%     \includegraphics[width=\textwidth]{figures/overall.pdf}
%     \caption{Overall training/fine-tuning procedure of \proposed{}}
%     \label{fig:three_phases}
% \end{subfigure}
% \hfill
% \begin{subfigure}{0.49\textwidth}
%     \includegraphics[width=\textwidth]{figures/fullattn.pdf}
%     \caption{Dense MHA operation}
%     \label{fig:fullMHA}
% \end{subfigure}
% \hfill
% \begin{subfigure}{0.49\textwidth}
%     \includegraphics[width=\textwidth]{figures/sparseattn.pdf}
%     \caption{Sparse MHA operation}
%     \label{fig:sparseMHA}
% \end{subfigure}
% \caption{\proposed{} splits the overall training process into three phases: dense-attention training with dense MHA operation, sparsity pattern generation, and sparse-attention training with sparse MHA operation}
% \label{fig:overview}
% \end{figure*}

% This section analyzes the sparsity patterns in the MHA operation.
% In this section we analyze the MHA operation to recognize the sparsity patterns.

\begin{figure*}
\centering
\begin{subfigure}{0.95\textwidth}
    \includegraphics[width=\textwidth]{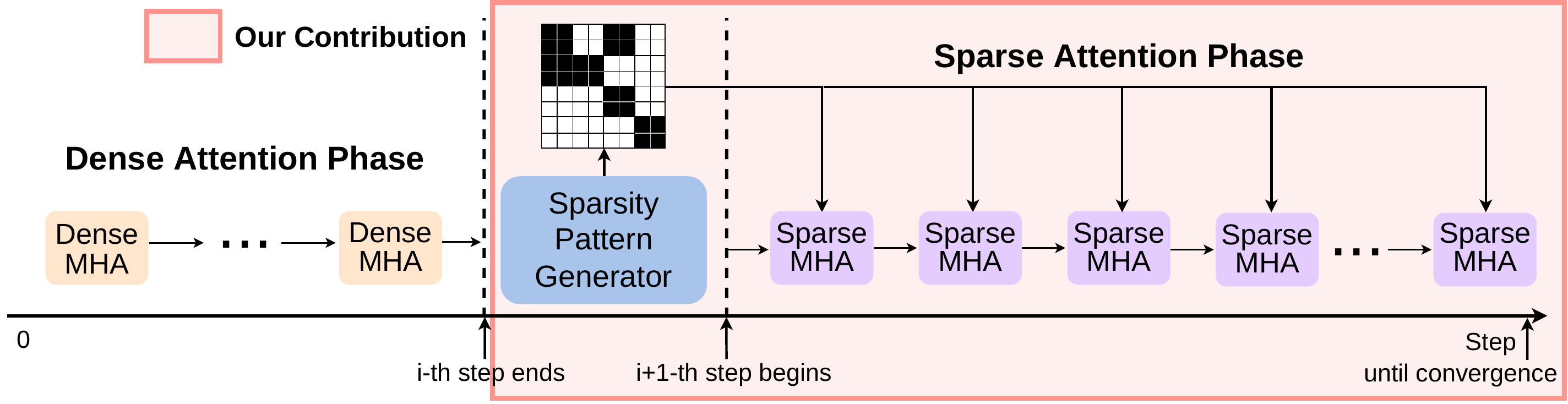}
    \caption{Overall training/fine-tuning procedure of \proposed{}}
    \label{fig:three_phases}
\end{subfigure}
\hfill
\begin{subfigure}{0.49\textwidth}
    \includegraphics[width=\textwidth]{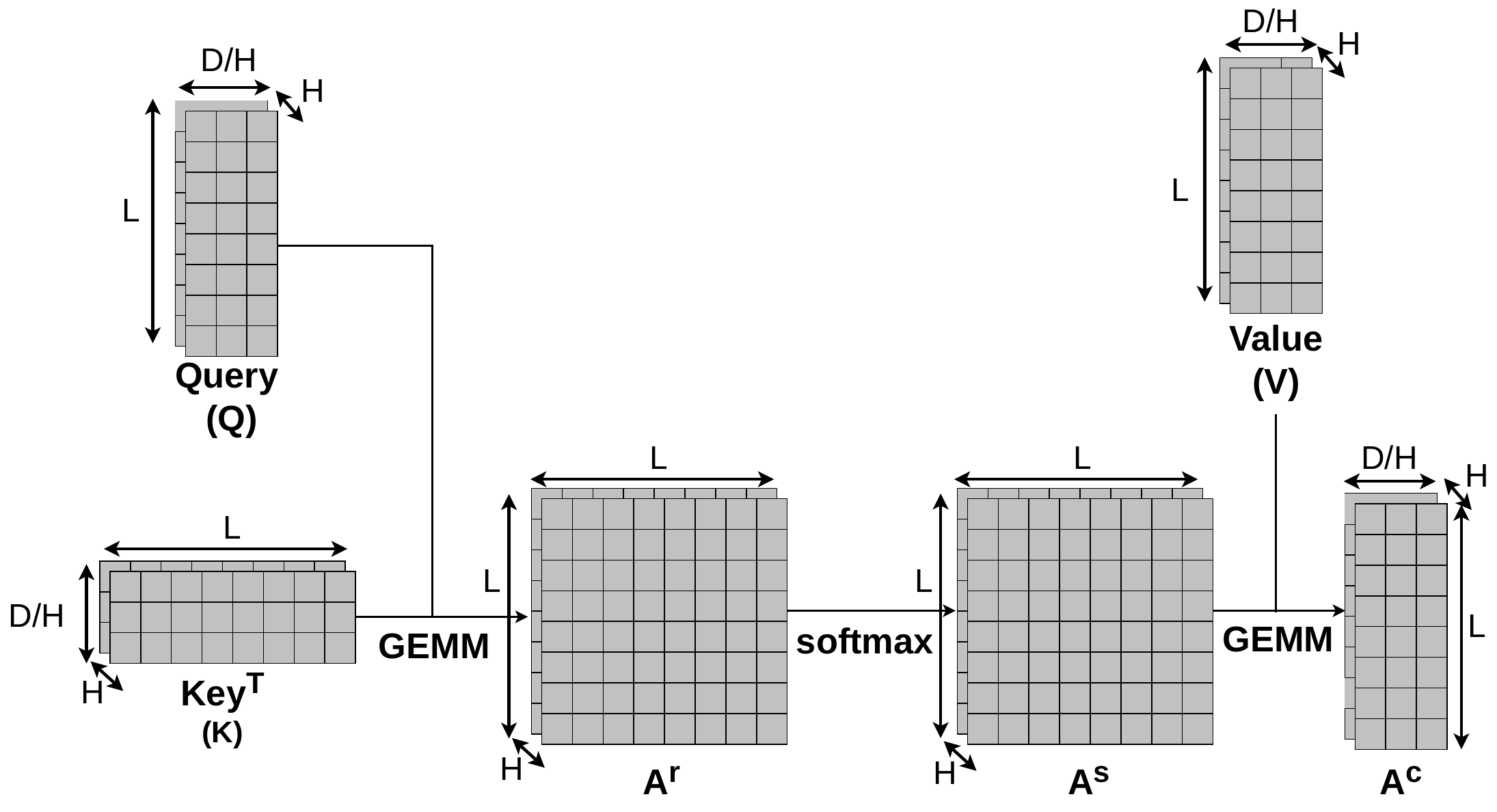}
    \caption{Dense MHA operation}
    \label{fig:fullMHA}
\end{subfigure}
\hfill
\begin{subfigure}{0.49\textwidth}
    \includegraphics[width=\textwidth]{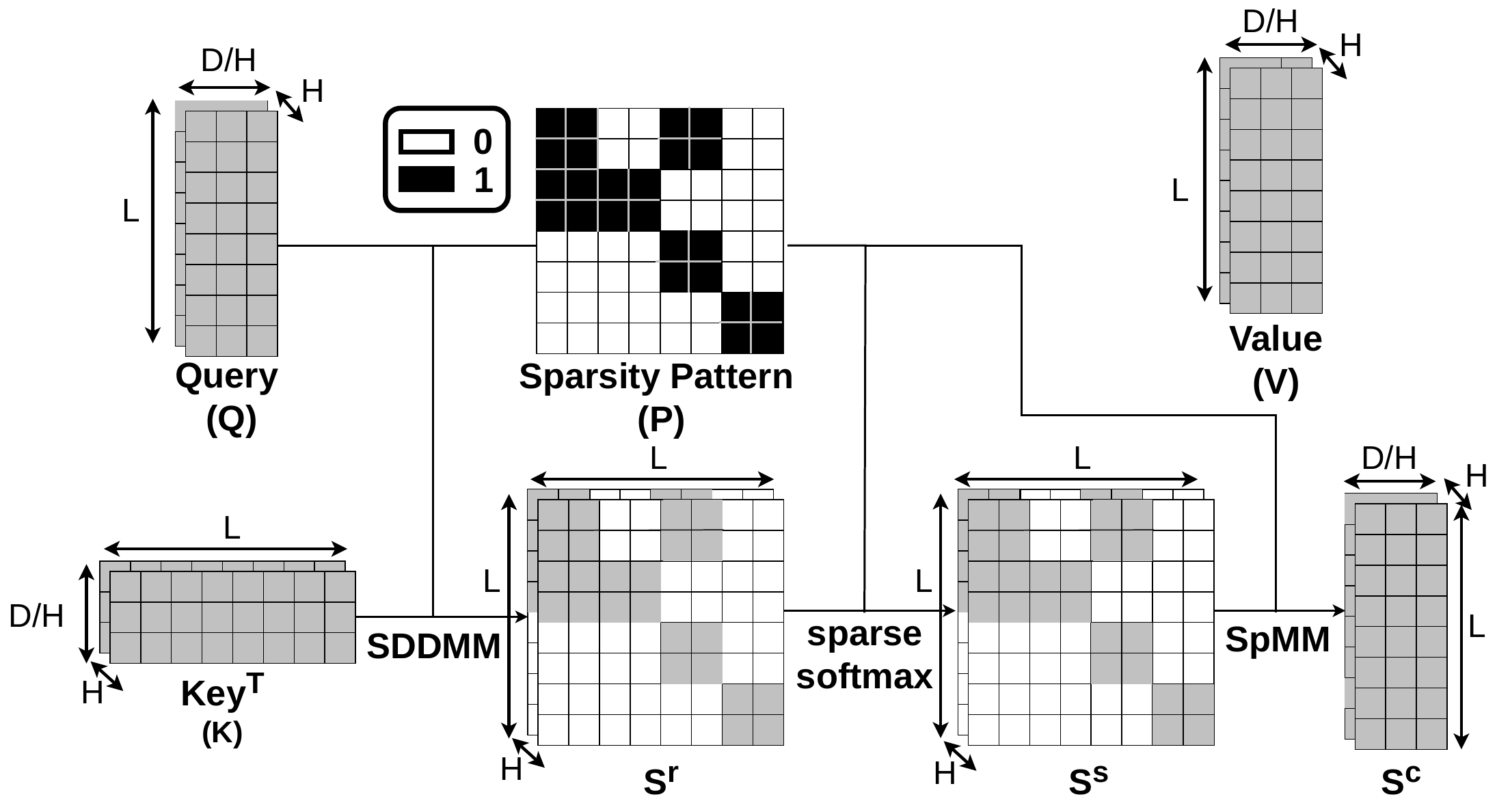}
    \caption{Sparse MHA operation}
    \label{fig:sparseMHA}
\end{subfigure}
\caption{\proposed{} splits the overall training process into three phases: dense-attention training with dense MHA operation, sparsity pattern generation, and sparse-attention training with sparse MHA operation}
\label{fig:overview}
\end{figure*}

% \rev{REVIEW 1: Instead, the authors should explicitly mention in motivation that "what’s important missing features they observed in prior works should be preserved" and "why increasing in model size from prior works causes a real and big problem."}

% \rev{REVIEW 2: What are the unique discoveries/motivations not found in previous work?}

% \gm{Emphasize layer-wise sparsification is required, head-wise.. pruning of MHA}

% In this section, we analyze the sparsity patterns in the MHA operation.
% In order to recognize the common sparsity patterns, we conducted fine-tuning experiments on the pretrained ViT \cite{dosovitskiy2020image} using the CIFAR-10 dataset.
In order to recognize the common sparsity patterns in MHA operation, we conducted experiments on the pretrained encoder-only Transformer \cite{dosovitskiy2020image}.
% Note that the pretrained ViT model is trained on a large ImageNet-21K dataset. 
Note that the pretrained model is trained on a large ImageNet-21K dataset.
Figure \ref{fig:motivation} illustrates the attention score matrices $A^s$ from different encoder layers during the first epoch of fine-tuning the model with the CIFAR-10 dataset.
Since the sparsity patterns of multiple attention score matrices within the same encoder layer typically show similar patterns, we averaged the attention score matrices across multiple heads in each encoder layer.
The results clearly show that most of the elements in $A^s$ are close to zero, indicating that only a few data points in the input sequence are correlated to each other and considered critical.
In practice, as the sparse pattern of matrix $A^s$ is maintained for each layer after training the model for a few epochs, a number of studies have shown that considering only the critical data points does not adversely affect the convergence of the model \cite{beltagy2020longformer,zaheer2020big,qiu2019blockwise}.
% The characteristics of sparsity patterns in the attention score matrices observed from our experiments are as follows.
The characteristics of attention score matrices $A^s$ observed from our experiments are as follows.

\betterparagraph{Shape of Sparse Pattern}
As shown in Figure \ref{fig:motivation}, the attention score matrices $A^s$ produced by the different encoder layers exhibit distinct sparsity patterns.
For example, in the first to eighth encoder layers, the diagonal elements have relatively large values, similar to a band matrix that stores nonzeros within the diagonal band of the matrix.
It is obvious that the MHA operation relies on the self-attention scheme and therefore, the resulting values of the dot-product between linearly transformed vectors for the same data points tend to be larger compared to the resulting outputs produced with different data points.
In addition to the diagonal sparsity pattern, some encoder layers, specifically layers 9, 10, 11 and 12, show a vertical sparsity pattern, with nonzeros mostly stored in specific columns. 
This vertical sparsity pattern emerges when the attention operation focuses on the similarity between all data points in $Q$ and particular data points in $V$.
% \gm{Based on these observations, applying the same fixed sparse pattern to all layers may lead to the loss of distinct vital features that need to be captured at each of different layers.}
In light of these observations, applying the same fixed sparse pattern to all layers may lead to the exclusion of unique essential features that need to be captured individually at different layers.
Hence, during the training of the Transformer model, to efficiently reduce the number of operations, it is crucial to consider layer-wise sparsification of the MHA based on the sparse pattern observed across different layers.
% Furthermore, the flexibility of changing the sparse pattern needs to be considered during training of the model. 
% Furthermore, generating domain-specific sparse pattern is required for various datasets and tasks.
Furthermore, the flexibility of changing the sparse pattern needs to be considered during the training of the model. Moreover, generating domain-specific sparse patterns is required for various datasets and tasks.

\betterparagraph{Degree of Sparsity}
% The number of nonzero elements in $A^s$ varies across encoder layers. 
Across different encoder layers, there exists variation not only in the shape of the sparse pattern but also in the number of nonzero elements in $A^s$.
% \gm{The number of nonzero elements in $A^s$ is irregular across different encoder layers.}
% \gm{The distribution of nonzero elements in $A^s$ is irregular across encoder layers.}
For example, layer 12 has a higher number of nonzero elements compared to layer 2, indicating that layer 12 extensively computes the MHA operation using a larger number of data points in the sequence.
% Therefore, by considering different degrees of sparsity, a layer-wise MHA computation can significantly reduce the total number of MHA operations and memory consumption across all encoder layers.
Hence, it is crucial to consider varying degrees of sparsity for every encoder layer.
This approach is essential for effectively reducing computational operations while preserving key features across distinct encoder layers.

\section{\proposed{}: Layer-Wise Sparse Attention in Transformer}
\label{sec:model}

\subsection{Overview of \proposed{}}
\label{subsec:overview}

% \rev{Add "layer-wise sparsification of MHA", "layer-wise MHA operation" as many as possible in Section 4 (the submitted paper does not include the term "layer-wise" in Section 4).}

% \gm{Add pseudo-code for overview of \proposed{}}
\begin{algorithm}[!ht]
\caption{Overall Training Procedure of \proposed{}}
\label{alg:overview}
% \footnotesize
% \scriptsize
\SetAlgoLined
\DontPrintSemicolon
\KwIn{$E$: input embedding vectors, $L$: length of input sequence, $D$: embedding size, $N$: the number of encoder layers,
% $\alpha$: transition condition value for frobenius distance difference
$\alpha$: threshold for transition
}

$transition \gets False$

\For{$i = 0$ \KwTo $num\_epochs-1$}{

\Comment*[r]{Dense MHA Phase}
\If{transition == False}{

\For{$n = 0$ \KwTo $N-1$}{

$O, A_{i}^s \gets MHA(E)$

$E \gets Feed\_Forward(O)$

}

\If{i > 1}{

% \Comment*[r]{\gm{Why we need both $distance_{i-1}$ and $distance_{i}$? Justify this in Section 4.1}}

$distance_{i-1} \gets frobenius(A_{i-2}^s,A_{i-1}^s)$ 

$distance_{i} \gets frobenius(A_{i-1}^s,A_{i}^s)$ 

\If{$\sqrt{(distance_{i-1}-distance_{i})^2} < \alpha$}{

$transition \gets True$

$P \gets generate\_pattern(A_{i}^s, t)$
}

}

}

%\Comment*[r]{\gm{Below if statement (i > 1 and ...) is not included (nested) in the outer if statement?}}

\Comment*[r]{Sparse MHA Phase}

\If{transition == True}{

\For{$n = 0$ \KwTo $N-1$}{

$O \gets sparseMHA(E, P_n)$

$E \gets Feed\_Forward(O)$

}

}

}
\end{algorithm}
% \input{algorithms/sparseMHA}

% \begin{figure*}
%   \includegraphics[width=\textwidth]{figures/overview.pdf}
%   \caption{Overall training procedure of \proposed{}. The sequence length ($L$), number of heads ($H$), embedding size ($D$) are set to 8, 2, and 6, respectively. \gm{Increase the size of "Creating Sparse Attention Pattern"}}
%   \label{fig:overview}
% \end{figure*}
% \index{figure}

% \begin{figure*}
% \centering
% \begin{subfigure}{0.95\textwidth}
%     \includegraphics[width=\textwidth]{figures/overall.pdf}
%     \caption{Overall training/fine-tuning procedure of \proposed{}}
%     \label{fig:three_phases}
% \end{subfigure}
% \hfill
% \begin{subfigure}{0.49\textwidth}
%     \includegraphics[width=\textwidth]{figures/fullattn.pdf}
%     \caption{Dense MHA operation}
%     \label{fig:fullMHA}
% \end{subfigure}
% \hfill
% \begin{subfigure}{0.49\textwidth}
%     \includegraphics[width=\textwidth]{figures/sparseattn.pdf}
%     \caption{Sparse MHA operation}
%     \label{fig:sparseMHA}
% \end{subfigure}
% \caption{\proposed{} splits the overall training process into three phases: dense-attention training with dense MHA operation, sparsity pattern generation, and sparse-attention training with sparse MHA operation}
% \label{fig:overview}
% \end{figure*}

\begin{figure*}
  \includegraphics[width=\textwidth]{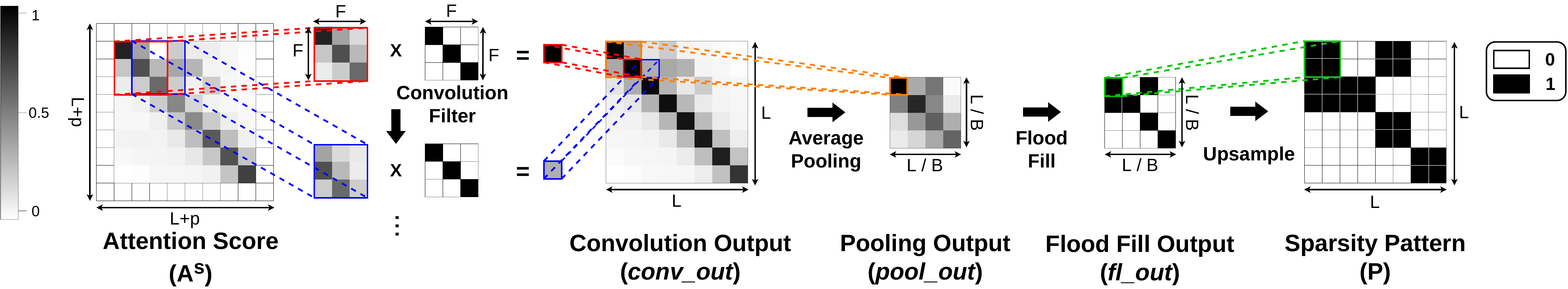}
  \caption{Overview of our convolutional flood filling method in the sparsity pattern generation/detection phase}
  \label{fig:patternoverview}
\end{figure*}
\index{figure}

% In this section, we present a high-level overview and details of our new \proposed{} model. 
% In this section, we present a high-level overview and details of our new \proposed{} model that overcomes the limitations of previous sparse attention approaches based on the major considerations described in Section \ref{sec:motivation}.
In this section, we provide a high-level overview and details of our new \proposed{} that dynamically sparsifies the MHA operation, incorporating the major considerations described in Section \ref{sec:motivation}. 
Note that our \proposed{} is capable of sparsifying the MHA operation either after training the model for a few steps or during the process of fine-tuning the model from the scratch using the pre-trained network.
% Note that our \proposed{} is capable of sparsifying the MHA operation during both training from the scratch/after few steps and fine-tuning with the pre-trained network.

% Based on the major considerations described in Section \ref{sec:motivation}, 
% Based on these experimental results, we propose a model that utilizes an algorithm aiming to capture the diagonal pattern of $A^s$ for each layer and sparsify elements that are considered less important. By doing so, a significant reduction in number of operations is achieved within the attention mechanism, leading to lower training time and memory usage.
% To overcome the limitations of previous sparse attention approaches described in Section \ref{subsec:related_work}, we propose \proposed{}, a dynamic sparsity-aware Transformer model that sparsifies the attention mechanism to reduce \bk{computation?}.
% To overcome the limitations of previous sparse attention approaches, we propose \proposed{}, a sparsity-aware dynamic Transformer model based on the main considerations described in Section \ref{sec:motivation}.
%sparse attention mechanism to reduce the model's training time and memory usage. 
%Sparsity can be achieved in deep learning models through various techniques, such as 'train and sparsify,' 'sparsify during training,' and 'sparse training'\cite{hoefler2021sparsity}. In this paper, we adopt the 'sparsify during training' approach, which involves sparsifying the model during training before the model fully converges.
As shown in Figure \ref{fig:three_phases}, we decouple the overall training process of \proposed{} into three phases: dense-attention training, sparsity pattern generation, and sparse-attention training.
% Figure \ref{fig:overview} shows the overall training process of \proposed{}, which can be divided into three phases: full-attention training, sparsity pattern generation, and sparse-attention training.
% As shown in Figure \ref{fig:overview}a, 
The dense-attention training follows the same training procedure as the original Transformer, without sparsifying the MHA operation.
During the dense-attention training phase, the original MHA, which includes two GEMM kernels and a dense softmax kernel (Figure \ref{fig:fullMHA}), is performed until the attention score matrix $A^{s}$ of each encoder layer exhibits a specific sparsity pattern.
% For each epoch in the full-attention phase, each MHA essentially performs two GEMM operations and one softmax operation. This original MHA continues to perform until the attention score matrix $A^{s}$ of each encoder layer exhibits a specific sparsity pattern.
% for a few training steps/epochs until the attention score matrix $A^{s}$ of each encoder layer exhibits a specific sparsity pattern.
% Here, each step refers to each process of updating the model parameters based on a mini-batch.
% Based on the validation accuracy during training, 
% If the \gm{XXX} condition is satisfied, 
% If the validation accuracy does not show improvement over several steps, 
% \gm{Need to add the details of conditions, add the equation of frobenius norm} 
% \gm{@BK: Write the equation of cost function (measuring the distance between two matrices) used in \proposed{}}
In order to determine the end of the dense-attention phase or the start of the sparse-attention phase, we measure the Frobenius distance between the attention score matrices $A^{s}$ produced in the previous step $i-1$ and the current step $i$ as defined in Equation \ref{eq:distance}. 

\begin{equation}
\label{eq:distance}
%distance_i = \sqrt{\sum (A^s_{i-1} - A^s_i)^2}
distance_i = \bigg | \sqrt{\sum (A^s_{i-1})^2} - \sqrt{\sum (A^s_{i})^2} \bigg | 
\end{equation}
% If there is no improvement in the validation accuracy over several steps, 
% If there is no significant difference in distance but there is improvement in the validation accuracy over several steps, 
%but there is improvement in the validation accuracy over several steps, 
If there is no significant difference in distance,
our \proposed{} ceases the dense-attention training phase and dynamically generates the sparsity pattern for each encoder layer based on our novel convolutional flood-filling scheme, which is described in Section \ref{subsec:sparsity_gen}.
Intuitively, it can be assumed that the attention score matrix is generalized when the values in the current attention score matrix are similar to those in the previous attention score matrix.
This indicates that it is ready to apply sparsification to the attention score matrix.
% \bk{The meaning of there is no significant difference bewteen the attention score matrices is that the attention score matrices are getting generalized and ready to apply the sparse pattern.}
% \gm{Need to describe/justify how the value of $\alpha$ is determined.}
Thereafter, \proposed{} proceeds with the sparse-attention training phase until convergence by adapting the sparsity pattern, as depicted in Figure \ref{fig:three_phases}.
Given the dense matrices $Q$ and $K$, along with the newly generated sparsity pattern matrix $P$, the SDDMM (Sampled Dense-Dense Matrix Multiplication) kernel is utilized to accelerate producing the sparsified attention score matrix $S^{r}$. Next, the sparse matrix $S^{r}$ is used to perform the optimized sparse softmax kernel, leveraging warp-level reduction for accelerating performance.
After applying the sparse softmax operation, since the attention score matrix $S^{s}$ remains sparse, we utilize SpMM (Sparse-Dense Matrix Multiplication) kernel to multiply the sparse matrix $S^{s}$ with the dense matrix $V$ (Figure \ref{fig:sparseMHA}).
% While performing the sparse-attention training phase, 
% the sparse MHA computation requires the SDDMM (Sampled Dense-Dense Matrix Multiplication) kernel, sparse softmax kernel, and SpMM (Sparse-Dense Matrix Multiplication) kernel (Figure \ref{fig:sparseMHA}).

% As the sparse training phase takes longer than full training phase, we optimize the softmax kernel.

% After completing the full-attention training phase, 

% The left part of the figure illustrates the training process of the model using the dense attention scheme from the original Transformer model. 
% This attention mechanism involves two GEMM operations and one softmax operation. This process repeats for a few steps (iterations). 
% A step refers to the process of updating the model's parameters once, based on a subset of the training dataset (called a "batch").
% After a few steps of training using the dense attention mechanism, we dynamically generate a sparsity pattern of the attention score matrix. This sparsity pattern is then adapted to the sparse attention mechanism.
% Since we have observed that the sparsity pattern of $A^s$ varies across layers, we generate a sparsity pattern for each layer.
% Lastly, the right side of Figure \ref{fig:overview} represents the remaining training process of a model using the sparse attention mechanism. This mechanism includes one SDDMM operation, one SpMM operation, and one sparse softmax operation.

% \subsection{Sparse MHA with Convolutional Flood Fill Algorithm}
\subsection{Sparsity Pattern Generation with Convolutional Flood Fill Algorithm}
\label{subsec:sparsity_gen}

\begin{algorithm}[!ht]
\caption{Implementation of generate\_pattern() function}
\label{alg:pattern}
% \footnotesize
\KwIn{$A^s \in \mathbb{R}^{L \times L}$, $t$ : threshold value for flood fill function}
\KwOut{$P$: sparsity pattern matrix}

$filter \gets \text{generate diagonal filter: } (F\times F)$

$conv\_out \gets \text{convolution}(A^s, filter)$

$pool\_out \gets \text{avgpool}(conv\_output,(B, B))$

$fl\_out \gets$ initialize with zeros : $(L/B \times L/B)$

\For{$i = 0$ \KwTo $L/B-1$}{
    $pool\_out, r, c, fl\_out, t \gets \text{flood\_fill}(pool\_out, 0, i, fl\_out, t)$
}

\For{$j = 0$ \KwTo $L/B-1$}{
    $pool\_out, r, c, fl\_out, t \gets \text{flood\_fill}(pool\_out, j, 0, fl\_out, t)$
}

\For{$k = 0$ \KwTo $L/B-1$}{
    $fl\_out[k,k] \gets 1$
}

$P \gets \text{upsampling}(fl\_out,(B,B))$
\end{algorithm}

% \begin{figure*}
%   \includegraphics[width=\textwidth]{figures/patternall.pdf}
%   \caption{Overview of convolutional flood-filling method in the sparsity pattern generation phase}
%   \label{fig:patternoverview}
% \end{figure*}
% \index{figure}

% \begin{figure}[!ht]
% \centering
% \includegraphics[width=0.49\textwidth]{figures/convpattern.pdf}
% \caption{Searching Pattern with convolution filter}
% \label{fig:convpattern}
% \end{figure}
% \index{figure}

To precisely detect the sparsity patterns in the attention score matrix $A^{s}$ generated during dense-attention training, we develop a new convolutional flood fill algorithm that extensively explores the shape, degree and locality of sparsity patterns for each encoder layer.
% systemically
% We propose our new convolutional flood fill algorithm that dynamically captures the sparsity patterns in the attention score matrix and generates sparse patterns of the attention score matrix. 
%First, it assumes the significance of the diagonal-shaped pattern. 
%\bk{The first assumption is that a diagonal-shaped pattern in the attention score matrix holds significance.
%And the second assumption is that neighboring data points within a specific area in the attention score matrix have strong relationships.
%These assumptions have been verified through the motivation experiment.}
% In the sparsity pattern generation phase, an initial step is to apply a diagonal convolution filter to $A^{s}$ in order to detect the presence of a diagonal shape of pattern in $A^{s}$, as shown in Figure \ref{fig:patternoverview}.
Algorithm \ref{alg:pattern} shows the pseudo-code for generating the sparsity pattern in the attention score matrix $A^{s}$.
An initial step is to apply a diagonal convolution filter to $A^{s}$ (line 1 in Algorithm \ref{alg:pattern}) in order to detect the shape of sparsity pattern in $A^{s}$, as shown in Figure \ref{fig:patternoverview}.
% Utilizing a diagonal convolution filter that stores value 1 for the diagonal element 
% To highlight the diagonal pattern in the attention score matrix $A^s$, we utilize a convolution filter that is assigned a value of 1 to the main diagonal elements and 0 to others. 
% By applying a diagonal convolution, it is possible to focus on not only diagonal elements (pattern) but also outside of the diagonal elements.
% If the diagonal elements in the $A^{s}$ have larger values than others, applying a diagonal convolution provides the convolution output where the diagonal elements are more emphasized while maintaining the values of off-diagonal elements.
If the diagonal elements in $A^{s}$ have larger values compared to the others, applying a diagonal convolution filter increases the values of the diagonal elements in the convolution output ($conv\_out$), leading to the emergence of a diagonal sparsity pattern.
Otherwise, if the off-diagonal elements, especially the vertical ones, in $A^{s}$ have larger values compared to the others, applying a diagonal convolution filter results in a vertical sparsity pattern in $conv\_out$ matrix.
% while preserving the values of the off-diagonal elements. 
% This makes 
% we can emphasize the elements within a diagonal pattern, while remaining other important values.
% dynamically capture the 
% (nonzero?) elements within the diagonal pattern. 
% \gm{why CF is applied for detecting elements in the diagonal pattern?}
%, in accordance with the first assumption.
% To highlight the diagonal pattern in the attention score matrix $A^s$, we utilize a filter that is assigned a value of 1 to the main diagonal elements and 0 to others. 
% By applying this type of filter, we can emphasize the elements within a diagonal pattern, while remaining other important values.
%, as shown in the first line of Equation \ref{eq:convfilter}.
%while retaining other important values.
%The filter design is motivated by the fact that the convolution operation performs element-wise multiplication and addition with the filter, as shown in the first line of equation \ref{eq:convfilter}. 
%As a result, the filter produces an output that emphasizes the diagonal pattern while retaining other important values. 
% To make the same size of matrices between $A^{s}$ and the convolution output, ($L \times L$), we add the amount of p zero-padding to the $A^{s}$ while performing the convolution operation.
Note that in order to ensure that the attention score matrix $A^{s}$ and the convolution output ($conv\_out$) have the same size ($L \times L$), we adopt zero-padding to $A^{s}$ during computing the convolution operation defined in Equation \ref{eq:convfilter}.
% employ a stride size of 1 for the convolution operation and apply zero-padding to maintain the size of the input matrix ($L \times L$). 
% we employ a stride size of 1 for the convolution operation and apply zero-padding to maintain the size of the input matrix ($L \times L$). 
%This results in an output that enhances the diagonal-shaped pattern while preserving the information from other crucial patterns or important values. 
% As illustrated in Figure \ref{fig:patternoverview}, the values in the diagonal patterns of the Convolution Output became larger compared to the values before the convolution operation. 
% Consequently, the algorithm can more effectively identify the elements within the diagonal pattern when generating sparsity pattern later on.
\begin{equation}
\label{eq:convfilter}
conv\_out(i,j) = \sum_{f=1}^{F} A^s(i+f,j+f) \times filter(f,f)
\end{equation}
%Based on the second assumption mentioned earlier, which states that
%%%%%%%%%
% As seen in Figure \ref{fig:motivation}, the adjacent data points within a specific region can be closely related. Therefore, to compute the attention score for critical data points and their surrounding data points, \proposed{} forms blocks of data points based on the output from the flood fill algorithm.
%%%%%%%%%

% In the process of generating the block sparsity pattern matrix $P$, 
After generating the $conv\_out$ through a diagonal convolution operation,
our algorithm performs average pooling on the $conv\_out$ using a kernel/block of size ($B\times B$) as defined in Equation \ref{eq:pooling} (line 2 in Algorithm \ref{alg:pattern}).
\begin{equation}
\label{eq:pooling}
pool\_out \left( \frac{i}{B},\frac{j}{B} \right) = \frac{1}{B^2} \sum_{p=1}^{B} \sum_{q=1}^{B} conv\_out(i+p,j+q)
\end{equation}
Instead of analyzing the sparsity pattern of $A^{s}$ element by element, applying average pooling enables capturing block sparsity pattern, which takes into account both the critical data points and their surrounding data points.
% Then the size of the average pooling output ($pool\_out$) is smaller than that of the attention score matrix $A^{s}$, indicating that $pool\_out$ can be considered as the abstract \gm{sparsity} representation of $A^{s}$.
Hence, since the output of average pooling ($pool\_out$) has a smaller size ($L/B\times L/B$) compared to the attention score matrix $A^{s}$, $pool\_out$ can be considered as an abstract sparsity representation of $A^{s}$ in blocks.
% as a abstract representation of the sparsity pattern of $A^{s}$ in blocks. 
% as the block-wise abstract sparsity representation of $A^{s}$.
% maximize the data locality, our \proposed{} forms a block of elements 
% Blocked sparsity pattern
%%%%%%%%
% As the adjacent data points within a specific area can be closely related, we explore and generate sparsity patterns at the block level instead of individual element level.
%Instead of exploring relevant elements for patterns individually, our algorithm explores patterns within blocks of elements. 
%This means that if one block is considered important, then all elements within that block can be considered important. 
% Therefore, our algorithm applies average pooling using a kernel of size ($B\times B$), which corresponds to the size of one block. 
% This produces a block-shaped pattern of $A^s$ with a size of ($L/B\times L/B$), abstracted in a block level.
%Figure \ref{fig:patternoverview} illustrates this process, highlighting the importance of the diagonal pattern through the blocked shaped.
% \begin{equation}
% \label{eq:convfilter}
% \begin{aligned}
% &conv\_out(a,b) = \Sigma_{i}^{B} A^s(a+i,b+i) \times filter(i,i) \\
% &pool\_out(a/B,b/B) = \frac{1}{B^2} \Sigma_{i}^{B} \Sigma_{j}^{B} conv\_out(a+i,b+i)
% \end{aligned}
% \end{equation}

\begin{algorithm}[!ht]
\caption{Implementation of flood\_fill() function}
\label{alg:floodfilling}
% \footnotesize
\KwIn{$pool\_out \in \mathbb{R}^{L/B \times L/B}$, $r$: current row index, $c$: current column index, $fl\_out\in \mathbb{R}^{L/B \times L/B}$, $t$: threshold value }

\If{$(r+1 == L/B)$ \KwOr $(c+1 == L/B)$}{
\Return $pool\_out$, $r$, $c$, $fl\_out$, $t$ 
}

$m \gets \text{max}(pool\_out[r+1][c], pool\_out[r][c+1], pool\_out[r+1][c+1])$

\If{$pool\_out[r+1][c] == m $ \KwAnd $ fl\_out[r+1][c] == 0$}{

\If{$pool\_out[r+1][c] > t$}{

$fl\_out[r+1][c] \gets 1$

}

$pool\_out, r, c, fl\_out, t \gets \text{flood\_fill}(pool\_out, r+1, c, fl\_out, t)$ 

}

\If{$pool\_out[r][c+1] == m $ \KwAnd $ fl\_out[r][c+1] == 0$}{

\If{$pool\_out[r][c+1] > t$}{

$fl\_out[r][c+1] \gets 1$

}

$pool\_out, r, c, fl\_out, t \gets \text{flood\_fill}(pool\_out, r, c+1, fl\_out, t)$ 

}

\If{$pool\_out[r+1][c+1] == m $ \KwAnd $ fl\_out[r+1][c+1] == 0$}{

\If{$pool\_out[r+1][c+1] > t$}{

$fl\_out[r+1][c+1] \gets 1$

}

$pool\_out, r, c, fl\_out, t \gets \text{flood\_fill}(pool\_out, r+1, c+1, fl\_out, t)$ 

}

\Return $pool\_out$, $r$, $c$, $fl\_out$, $t$ 
\end{algorithm}

\begin{figure}[!ht]
\centering
\includegraphics[width=0.49\textwidth]{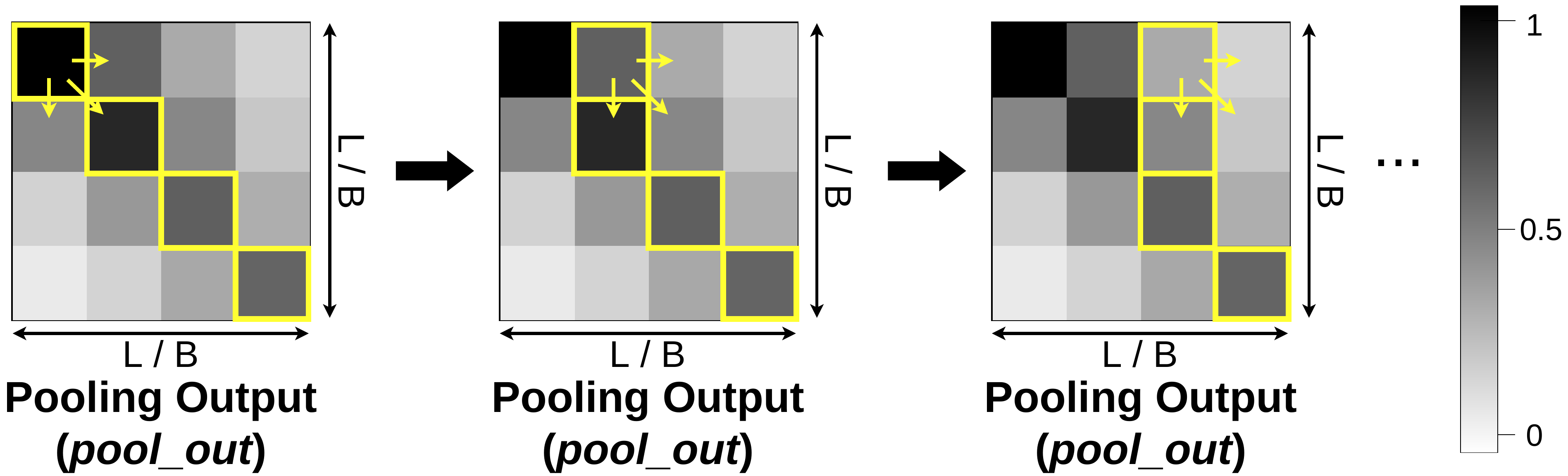}
\caption{Walk-through example of our flood fill algorithm
% \gm{We need to clearly show the walk-through example describing how the fl\_out in Figure 3 is generated given pool\_out in Figure 3}
}
% \caption{Exploring sparsity pattern in the Attention Score matrix with flood fill algorithm}
\label{fig:floodfilling}
\end{figure}

% Flood Filling (also considering number of nonzero elements in A^{s})
% \gm{(blocks of elements in $conv\_out$)}
% To dynamically explore the critical elements in the $pool\_out$, we utilize the flood fill algorithm described in Section \ref{subsec:flood_fill}.
% \rev{(our \proposed{} finds the sparsity pattern in the $pool\_out$ inspired by flood fill algorithm)}
%The flood fill algorithm is a method of determining the area connected to a given point in a multi-dimensional array or bitmap image through coloring.
% To dynamically explore the critical elements in the $pool\_out$ and generate $fl\_out$, we develop a new algorithm inspired by the flood fill algorithm.
In order to dynamically explore the crucial elements in the $pool\_out$ and generate $fl\_out$, we develop a novel algorithm inspired by the flood-fill algorithm.
% \rev{We develop a new algorithm for generating a sparsity pattern matrix inspired by the flood fill algorithm.}
By adapting the flood filling scheme, our \proposed{} is able to precisely analyze the connectivity between significant elements in $pool\_out$, while also considering the number of critical nonzero elements.
% Furthermore, the flood filling scheme allows our algorithm to consider the number of critical nonzero elements.
Algorithm \ref{alg:floodfilling} shows the pseudo-code for our flood fill-based algorithm recursively executed in Algorithm \ref{alg:pattern}.
Unlike the traditional flood fill algorithm, which compares all neighbors of a current element to find the element with the largest value, our new algorithm only compares to the neighboring elements on the right, below, and diagonally below, as shown in Figure \ref{fig:floodfilling}. 
In the process of capturing the pattern, it is necessary to sequentially follow the important features starting from the top-left corner to the bottom of the matrix.
Hence, comparing the current element with the elements to its right or below checks whether the neighbors of the current element are relevant.
If the neighboring elements are not relevant to the current element, our algorithm moves to the element diagonally below to continue comparing the neighbors.

% \gm{Need to justify the reason why only three elements are examined. Why not considering all eight elements?}
% \bk{The reason why we compare with elements on the right, below, and diagonally below is that it is necessary to capture the pattern sequentially as we are dealing with sequential data. From the top of the matrix, we sequentially follows the important features until it reaches to the bottom. To be specific, comparing with the elements on the right or below means that if the neighbors of the current element are relevant or not. And comparing with diagonally below element means that neighboring elements of current element is not relevant enough so that we would move to the next element to compare with one's neighbors.}
\index{figure}

% We first compare the values in the right, down, and diagonally below directions to find the largest value, $m$ (line 3). Then, we check whether $m$ is greater than a predefined threshold $t$. 
% Threshold $t$ is determined through calculating the $\alpha\%$ quantile of $pool\_out$. When calculating a quantile, the data is first sorted in ascending order. The $\alpha\%$ quantile represents the value above which $\alpha\%$ of the data in the sorted data falls.

% \rev{We need to show effectiveness (flexibility) of our algorithm (sparsity pattern generator) using text (in Section 4)}

The element at the first row and first column of the $pool\_out$ is used as the starting point (lines 4 and 6 in Algorithm \ref{alg:pattern}).
% Then the algorithm starts to seek the most \gm{similar? relevant? large value? important?} element among its neighboring elements to the right, below, and diagonally below. 
Then the algorithm compares the values of elements to its right, below, and diagonally below, extracting the element with the largest value, $m$ (line 3 in Algorithm \ref{alg:floodfilling}). When the value of $m$ is greater than a predefined threshold $t$ (line 5, 9, and 13 in Algorithm \ref{alg:floodfilling}), we determine the corresponding element as the critical element and assign a value of 1 to the corresponding element in $fl\_out$ (lines 6, 10 and 14 in Algorithm \ref{alg:floodfilling}).
Here, the threshold $t$, is determined by calculating the $\alpha\%$ quantile of $pool\_out$.
% When calculating a quantile, the data is first sorted in ascending order. The $\alpha\%$ quantile represents the value above which $\alpha\%$ of the data in the sorted data falls.
Hence, even if one of the neighbors is selected as a potential critical point, it may be classified as a non-critical point based on the threshold $t$. By utilizing the threshold $t$, we ensure that the selected critical elements have values that are sufficiently large.
After determining the critical element, the algorithm recursively compares the values of elements to the right, below, and diagonally below the critical element while avoiding duplicate comparisons with elements that have already been selected. (second condition of lines 4, 8, and 12).
% Once the critical element is selected, the algorithm searches for its \gm{closest} neighbor in order to establish links between critical elements.
% This process is repeated until the selected element reaches either the end of a row or the end of a column in the matrix $pool\_out$.
This process is repeated until the selected critical element reaches the end of a row or column in the matrix $pool\_out$ (line 1 in Algorithm \ref{alg:floodfilling}).
% either the end of a row or the end of a column in the matrix $pool\_out$.
% the most similar
Afterward, as shown in the middle of Figure \ref{fig:floodfilling}, the element at the first row and second column serves as the next starting point for recursively analyzing the connectivity of critical elements. 
Therefore, considering each element in $pool\_out$ as a seed point allows for the exploration of the connectivity of critical elements across all elements in $pool\_out$.
Since the attention score between the same data points in the sequence tends to be large for most of the encoder layers, we initially assign a value of 1 to the diagonal elements in $fl\_out$.

The final output of our flood fill algorithm ($fl\_out$) can be seen as a more explicit sparsity pattern captured from $pool\_out$ and can also be considered as the compressed block-level sparsity pattern of $A^{s}$.
To utilize the sparse pattern of $fl\_out$ in the sparse MHA operation, the size of the matrix $fl\_out$ needs to be the same as the size of the attention score matrix $A^s$. 
Therefore, we upsample the $fl\_out$ using nearest-neighbor interpolation (line 11 in Algorithm \ref{alg:pattern}), resulting in each nonzero element in $fl\_out$ forming a block of nonzero elements in the final sparsity pattern matrix ($P$), as shown on the right of Figure \ref{fig:patternoverview}.
% Utilizing a block sparse matrix can improve the quality of the model as the both the critical elements and its surrounding elements are used for MHA operation. 
Utilizing a block sparse matrix can improve the model's quality since it incorporates not only the critical elements but also their surrounding elements for MHA operation.
Moreover, an optimized blocked matrix multiplication further enhances performance through improved data locality.
In the blocked sparse matrix $P$, elements that require calculation in the attention score matrix are set to 1, while those that do not require calculation are set to 0.
% represents a block in the sparsity pattern ($P$), as shown in Figure \ref{fig:patternoverview}. 
% For example, if the element in the first row and first column of the Flood Fill Output ($fl\_out$) has a value of 1, it indicates that the elements in the rows from the first row to the B-th row and columns from the first column to the B-th column in the sparsity pattern ($P$) will have a value of 1.
%\bk{We will demonstrate the impact of creating a block-level attention sparse pattern in the experiments section.}
%$The meaning of a value of 1 in the sparsity pattern is that, when processing the sparse attention mechanism, only the positions where a value is 1 in the sparsity pattern will be performed.
% The value of 1 in the sparsity pattern indicates that, during the sparse attention, only the positions corresponding to 1 in the sparsity pattern will be performed.
% The value of 1 in the block sparsity pattern matrix $P$ indicates that,
Finally, during the sparse MHA, only the elements of $A^{s}$ that have the same indices as the elements with a value of 1 in the block sparsity pattern matrix $P$ will be computed.

\subsection{Acceleration of Sparse MHA Implementation on GPUs}
% \rev{remove "Backward Propagation of Sparse MHA" and "Sparse Softmax Kernel for Backward Propagation" paragraphs.}
In this sub-section, we provide details of our parallel sparse MHA implementation on GPUs.
% After generating the sparsity pattern matrix $P$ using our convolutional flood fill algorithm, 
Prior to the sparse-attention phase and sparsity pattern generation, we train the model with the dense MHA for several steps. 
During the dense-attention phase, our \proposed{} implementation uses the NVIDIA cuBLAS library with the tensor cores on GPUs, such as cublasGemmStridedBatchedEx(), to accelerate the dense MHA operation, which involves the multiplication of dense matrices $Q$ and $K$, as well as the multiplication of dense matrices $A^s$ and $V$.

\begin{algorithm}[!ht]
\caption{GPU Implementation of sparseMHA() on host}
\label{alg:sparseMHA}
% \footnotesize
% \scriptsize
\SetAlgoLined
\DontPrintSemicolon
\KwIn{$E$: a set of embedding vectors for an input sequence, $L$: length of input sequence , $D$: embedding size for each data point, $H$: the number of heads, $W^{Q}$, $W^{K}$, $W^{V}$, $W^{O}$: weight parameters, $P$: sparsity pattern matrix}

$X \gets LayerNorm(E)$

$Q \gets X\times W^{Q}, K \gets X\times W^{K}, V \gets X\times W^{V}$

% $Q_{0,...,H-1}$, $K_{0,...,H-1}$, $V_{0,...,H-1}$ $\gets \text{Splitting each $Q$, $K$, and $V$ into $H$ attention heads}$

$Q_{0,...,H-1}$, $K_{0,...,H-1}$, $V_{0,...,H-1}$ $\gets split(Q, K, V)$

\ParallelFor{$i = 0$ \KwTo $H-1$}{

    $S_{i}^{r} \gets cusparseSDDMM(Q_i, K_{i}^T, P)$

    $S_{i}^{s} \gets SparseSoftmax(S_{i}^{r}, P)$

    $S_{i}^{c} \gets cusparseSpMM(S_{i}^{s} ,V_i, P)$

}

$S \gets concatenate (S^{c}_{0}, ... , S^{c}_{H-1})$

$O \gets dropout(S \times W^{O}) + E$

\end{algorithm}
% \betterparagraph{Forward Propagation of Sparse MHA}
\betterparagraph{Sparse MHA}
% Algorithm \ref{alg:spattnfor} shows the pseudo-code for the forward propagation of sparse MHA. 
Algorithm \ref{alg:sparseMHA} shows the pseudo-code for the forward propagation of sparse MHA.
To accelerate the sparse MHA operation, we utilize NVIDIA cuSPARSE libraries such as cuspraseSDDMM and cusparseSpMM. Additionally, we optimize the softmax function to account for sparsity in the sparse raw attention score matrix.
% $S^{r}$.
% optimize our GPU implementation of sparse MHA.
% \proposed{} utilizes sparse attention mechanism with a sparsity pattern generated using the convolutional flood fill algorithm. 
% Therefore, we have implemented the optimized sparse attention mechanism as follows. 
%and adapts it to sparse attention mechanism.
%, leading to reduced training time and memory comsumption.

\begin{equation}
\label{eq:sparseattention}
S = softmax\left(\frac{(P > 0) \odot (Q \times K^T)}{\sqrt{(D/H)}}\right) \times V
\end{equation}
As shown in Equation \ref{eq:sparseattention}, the sparse MHA produces a sparse attention matrix $S$ by replacing the GEMM operation in line 7 of Algorithm \ref{alg:encoder} with the SDDMM operation using cusparseSDDMM (line 5 in Algorithm \ref{alg:sparseMHA}).
The SDDMM kernel computes the product of two dense input matrices $Q$ and $K$, and the resulting matrix is then subjected to a Hadamard product (element-wise multiplication) using a sampled sparse matrix $P$.
% at every nonzero position of a sampled sparse matrix $P$. 
In Equation \ref{eq:sparseattention}, $(P > 0)$ indicates that only the indices of nonzero elements in $P$ are computed in the result matrix while multiplying $Q$ and $K$, and $\odot$ denotes element-wise multiplication.
% takes two dense input matrices $Q$ and $K$ and a sparse matrix $P$ and computes a matrix multiplication of two dense matrices at every nonzero position of a sampled sparse matrix $P$.
% performing an exclusive matrix multiplication ($\odot$) using selected elements from $Q$ and $K$.
% In the sparse MHA operation, the GEMM operation in line 7 of Algorithm \ref{alg:encoder} is replaced with the SDDMM operation.
% where $\odot$ denotes an \gm{exclusive} matrix multiplication on the non-zero values.
Therefore, the SDDMM operation produces the sparsified raw attention matrix $S^{r}=(P>0) \odot (Q \times K^{T})$, which stores only a small number of nonzero elements.
% based on the indices of non-zero elements in the sparse pattern matrix $P$.
% In the sparse matrix $P$, elements that require calculation in the attention matrix are set to 1 (non-zero value), while those that do not require calculation are set to 0. 
To efficiently compute $S^{s}$ by performing the softmax operation with the sparse matrix $S^{r}$, we implement the sparse softmax function (line 6 in Algorithm \ref{alg:sparseMHA}). This function computes the probability distribution for only the nonzero values in $S^{r}$, instead of applying the standard softmax function used in line 8 of Algorithm \ref{alg:encoder}.
Thereafter, as $S^{s}$ remains a sparse matrix, we perform the SpMM operation using cusparseSpMM to compute the product of a sparse matrix $S^{s}$ and a dense matrix $V$ (line 7 in Algorithm \ref{alg:sparseMHA}).
% First, sparse MHA replaces the standard matrix multiplication of $Q$ and $K$ in line 7 of Algorithm \ref{alg:encoder} with Sampled Dense Matrix Multiplication (SDDMM). SDDMM involves multiplying a sparse matrix with two dense matrices.
%Specifically, this operation multiplies each non-zero entry in the predefined sparse pattern with the corresponding row in the first dense matrix $Q$ with the corresponding column in the second dense matrix $K$. 
% The softmax operation in line 8 of Algorithm \ref{alg:encoder} is also replaced with sparse softmax. %The softmax operation originally calculates the probability distribution of all elements within each row of the input matrix. In contrast, sparse softmax 
% It calculates the probability distribution for only the non-zero values in each row of the sparsity pattern $P$. 
% Lastly, the matrix multiplication of $A^s$ and $V$ in line 9 of Algorithm \ref{alg:encoder} is replaced with Sparse Matrix-Matrix Multiplication (SpMM). SpMM involves multiplying a dense matrix with one sparse matrix and one dense matrix. 
%Therefore, SpMM multiplies each non-zero entry in the sparse matrix $A^s$ with the corresponding column in the dense matrix $V$.
% To perform cursparseSDDMM, SparseSoftmax, and SpMM kernels with $P$, we convert the sparse matrix $P$ into the most commonly used Compressed Sparse Row (CSR) representation.
To perform cursparseSDDMM, SparseSoftmax, and SpMM kernels with $P$, we convert the sparse matrix $P$ into the most commonly used Compressed Sparse Row (CSR) format consisting of three data structures: $row\_ptr$, $col\_idx$ and $values$.

\begin{algorithm}[!ht]
\caption{SparseSoftmax() kernel for forward propagation on GPUs}
\label{alg:spsoftmax}
% \footnotesize
\KwIn{$S^r,P,scale$}

$warp\_id \gets \lfloor threadIdx.x / warp\_size \rfloor$

$lane \gets threadIdx.x  \%  warp\_size$

%$b\_cnt \gets cnt\_mat[warp\_id \% L]$

$b\_cnt \gets P.row\_ptr[(warp\_id + 1) \% L] - P.row\_ptr[warp\_id \% L]$

$b\_idx \gets \lfloor warp\_id/L \rfloor \times P.values + P.row\_ptr[warp\_id \% L]$

$max\gets -\infty$

$sum\gets 0$

\For{$k$ = $lane$ \KwTo $b\_cnt$-1 \KwBy $warp\_size$}{
    $S^r[b\_idx + k] \gets S^r[b\_idx + k] \times scale$
    
    \If{$max < S^r[b\_idx + k]$}{
    $max \gets S^r[b\_idx + k]$
    }
}

$max \gets warp\_reduce\_max(max)$

\For{$k$ = $lane$ \KwTo $b\_cnt$-1 \KwBy $warp\_size$}{
    $sum \gets sum + exp(S^r[b\_idx + k]-max)$
}
$sum \gets warp\_reduce\_sum(sum)$

$sum \gets sum +  exp(-max) \times (L - b\_cnt)$

\For{$k$ = $lane$ \KwTo $b\_cnt$-1 \KwBy $warp\_size$}{
    $S^r[b\_idx + k] \gets exp(S^r[b\_idx + k]-max) / sum$
}

\end{algorithm}
% \betterparagraph{Sparse Softmax Forward Kernel}
% \betterparagraph{Sparse Softmax Kernel for Forward Propagation}
\betterparagraph{Sparse Softmax Kernel}
Algorithm \ref{alg:spsoftmax} shows the pseudo-code for our SparseSoftmax() kernel.
% shown in Algorithm \ref{alg:spsoftmax}, 
In the SparseSoftmax() kernel, as the original softmax function operates in a row-wise manner, each warp is responsible for computing a single row. 
% This is because the softmax function in the original attention mechanism operates in a row-wise manner.
%$warp\_size$ is a constant with a value of 32. 
The variable $warp\_id$ indicates which row should be calculated on the current warp, and the variable $lane$ indicates the lane number within a single warp.
For each row, $b\_cnt$ indicates the number of elements that should compute, and $b\_idx$ indicates where the each row starts at $S^r$. 
%Adding $nnz \times blockIdx.x$ to $b\_idx$ is to find the corresponding row in multiple $S^r$ for multiple batches and heads. 
We initialize the variables $sum$ and $max$ to 0 and $-\infty$, respectively. Since there is no need to share these variables among different threads, the variables $sum$ and $max$ are initialized inside the kernel. To ensure numerical stability, we normalize the elements in each row by subtracting the maximum value found in each row.
%To find the maximum value in each row, a lane in a warp fetches and compares an element of a row iteratively skipping some elements nearby, so that other lane in same warp can handle the elements that were skipped(line 9 ~ 12). 
To find the maximum value in each row, a lane in a warp fetches and compares a subset of elements within that row (lines 9-11 in Algorithm \ref{alg:spsoftmax}).
%The function $warp\_reduce\_max$ allows a lane to exchange a value by using warp-level primitive then compares each other in reduction algorithm's manner. 
The function $warp\_reduce\_max()$ in line 11 enables a lane to exchange values using warp-level primitives and compare them with each other in a reduction manner.
After executing the $warp\_reduce\_max()$ function, each lane within a single warp receives the same maximum value that was passed by all the lanes.
%, which is the maximum value among all the values passed by the lanes.
%Then, we accumulate exponential of the elements that a single lane takes in charge. 
%Just like how we obtained the max value in line 13, we used warp-level primitive to aggregate sum value along all lanes in a single warp(line 16) to obtain sum value. 
We also use the warp-level primitive to aggregate sum value across all lanes within a single warp (line 14) and obtain the final value of $sum$.
%In line 17, we subtract max from the values we don't need to compute, assuming they are zero, to prevent the probability of non-zero values from becoming too large. 
% To prevent the probability of nonzero values from becoming very large, we add the values which are not required to compute to the value of $sum$ (line 17).
Finally, the computation of sparse softmax function is completed by dividing the elements in the row by the value of $sum$ variable.
%SparseSoftmax() kernel computes the max and sum values for the non-zero elements in the attention sparse pattern and writes the computed values back to the matrix for these elements only. This is in contrast to traditional softmax, which computes the length of the input sequence. 
Therefore, by only considering the nonzero elements in $S^{r}$, our SparseSoftmax() kernel greatly reduces the computational complexity of softmax operation.

\betterparagraph{Integration of GPU kernels with PyTorch}
% \rev{How are your GPU kernels integrated if you use a system like this (CUDA + PyTorch)?}
% \gm{Describe how to integrate CUDA kernels and PyTorch and how the backpropagation works in this way.}
%\lstinputlisting[language=Python,caption=\gm{Binding CUDA kernels with PyTorch Code}]{algorithms/example.py}
% \gm{I will edit this paragraph later}
% We integrate our CUDA kernels and CUDA libraries with PyTorch using the ctypes library (CDLL) in Python.
% In order to integrate our CUDA kernels and CUDA libraries with PyTorch, we use the ctypes library (CDLL) in Python.
To execute our optimized CUDA kernels and CUDA libraries for both forward and backward propagation within the PyTorch code, we integrate these kernels and libraries with PyTorch by employing the ctypes library (CDLL) in the Python environment.
Furthermore, our custom autograd function is used to process the backward propagation.
Before the training begins, we first compile our CUDA code into a shared object (.so) file.
Once compiled, the CUDA shared object file is loaded and then converted into a Python handle.
Subsequently, the Python handle is employed to invoke the CUDA kernels for both forward and backward propagations.
% Compiling is needed only once before or during the training process, and loading the CUDA shared object file and making it into a python handle are also needed only once before using it.
% We have each forward and backward CUDA kernels of SparseMHA, which can be called inside the PyTorch code.
% Compiling is needed only once before or during the training process, and loading the CUDA shared object file and making it into a python handle are also needed only once before using it. 
% We build our model with PyTorch custom autograd functions to use our own forward and backward functions; in custom autograd functions, we use the python handle made with the compiled shared object file to call the forward and backward kernels made with CUDA.

% \subsection{Analysis of Sparse MHA Operations}
% \subsection{Comparison Analysis of Sparse MHA Operations}
\subsection{Computational Complexity Analysis Comparison}
% \rev{Describe a reduction by an order of magnitude for the number of computations.}
By applying our sparsification scheme, the sparse raw attention matrix $S_{i}^{r}$ maintains a total of $C$ non-zero elements, whereas the original dense matrix $A_{i}^{r}$ maintains a total of $L \times L$ non-zero elements.
%$ncd(2D-1)+3ncd-L+D(2ncd-L)$ 
By using the sparse matrix $S_{i}^{r}$, the total number of operations required to compute $S$ is reduced to $2C(2D+1)-L(D+1)$, resulting in approximately $\frac{L^2}{C}\times$ reduction in operations compared to the original attention matrix $A$, which requires $2L^2(2D+1)-L(D+1)$ operations.
% generated by the original MHA approach. $L^{2}(2D-1)+L(3L-1)+LD(2L-1)$
In practice, for the real-world AAN dataset \cite{radev2013acl} ($L=4,096$ and $D=64$) with the number of $ncd=1,677,721$ (10\% of $L^2$), the total number of operations required for producing the original $A$ and the sparsified $S$ are $4,328,255,488$ and $432,585,778$, respectively. Therefore, sparse MHA requires approximately 10 times less operations than the original dense MHA during training. 

\section{Experimental Evaluation}
\label{sec:experiment}
This section provides both performance and quality assessments of our \proposed{} implementation. Our \proposed{} is compared with various state-of-the-art sparse Transformer models.

%\gm{Describe benchmarking machines and datasets used in the experiments. Add tables to show the machine configuration, statistics of datasets, and type of tasks/workloads.}

\begin{table}[ht]
\centering
\caption{Machine configuration}
\scalebox{0.9}{
\begin{tabular}{|c|c|}
\hline
Machine & Details                                                                                                          \\ \hline
CPU     & \begin{tabular}[c]{@{}c@{}}AMD Ryzen Threadripper PRO 5955WX\\ (16 cores and 32 threads, 128GB RAM)\end{tabular} \\ \hline
GPU     & \begin{tabular}[c]{@{}c@{}}NVIDIA RTX A5000\\ (24 GB Global Memory, 64 SMs, 6 MB L2 cache)\end{tabular}          \\ \hline
\end{tabular}}
\label{tab:machine}
\end{table}

\betterparagraph{Benchmarking Machines}
The detailed configuration of the benchmarking machines used for the evaluations is shown in Table \ref{tab:machine}. All the experiments were run on four NVIDIA RTX A5000 GPUs. 

\betterparagraph{Datasets and Tasks}
We evaluated our model on the Long Range Arena (LRA) \cite{tay2020long}, which is widely used for evaluating the effectiveness of efficient transformers for long sequences. 
% The LRA consists of five sequence classification tasks. Among these tasks, 
% LRA evaluations on the image classification, ListOps and document retrieval tasks.
We conducted three sequence classification tasks as part of the LRA evaluations, including image classification, ListOps and document retrieval tasks.
For all experiments, we measured classification accuracy to compare the quality of the compared models.
% , which measures the percentage of correctly classified data samples, as the evaluation metric.
The three tasks, along with the three datasets used in our evaluations, are as follows:
\begin{itemize}
    \item \textbf{Image classification}: This task classifies images into 10 different classes using the CIFAR-10 dataset \cite{krizhevsky2009learning}. Each image in the dataset has a size of 32x32 pixels, resulting in a sequence length of 1,024 by considering each pixel as a data point.
    \item \textbf{ListOps}: This task involves classifying the answer within the range of 0 to 9, given input equations expressed as a sequence of numbers and mathematical operators. The dataset provided from Nangia et al. \cite{nangia2018listops} is used for evaluating this task. The maximum sequence length is 2,048.
    \item \textbf{Document retrieval}: This task classifies whether two given documents are related or not. The AAN dataset \cite{radev2013acl} is used for this task, and the maximum sequence length is is 4,096.
\end{itemize}

% We utilized the LRA evaluation reimplemented in PyTorch by Xiong et al.\cite{xiong2021nystromformer}.

\betterparagraph{Models Compared}
We compared our \proposed{} with the original encoder-only Transformer and two state-of-the-art efficient sparse Transformers.
\begin{itemize}
    \item \textbf{Original encoder-only Transformer} \cite{vaswani2017attention}: This implementation is based on the original Transformer architecture and performs the original dense MHA operation during entire training.
    \item \textbf{BigBird} \cite{zaheer2020big}: This model incorporates sparse attention mechanisms, including sliding window attention, global attention, and random attention. It is evaluated using a block size of 64 and 3 random blocks.
    \item \textbf{Reformer} \cite{kitaev2020reformer}: This model utilizes a sparse attention mechanism based on locality-sensitive hashing algorithm. It is evaluated using a bucket size of 32 and 2 hashes.
    %We tried bigger size of bucket and more hashes but it only takes more time and less increment on classification accuracy.
    \item \textbf{\proposed{}-C}: This model is the variation of our \proposed{} model without applying the flood filling scheme. Instead, it selects the top $\alpha\%$ of block elements from the results of the convolution filter and average pooling ($pool\_out$), which enables adjustment of the sparsity ratio when generating the $P$.
    % \gm{which allows for adjusting the sparsity of $A^{s}$.}
    %Therefore, we set the density to 5\% for image classification, 4\% for ListOps and 3\% for document retrieval, respectively.
    % \item \textbf{\proposed{}-F}: This model is the variant of our \proposed{} without using a convolution filter. Instead, it directly applies the flood fill algorithm to the latest $A^{s}$ after performing average pooling.
    \item \textbf{\proposed{}-F}: This model is the variant of our \proposed{} without using a convolution filter. The model directly applies the flood-fill algorithm immediately after the average pooling operation without applying a convolution filter.
    \item \textbf{\proposed{}-CF}: This implementation incorporates both the convolution filter and the flood fill-based scheme while generating sparsity pattern.
\end{itemize}

% \betterparagraph{Baselines}
% Aside from the original Transformer model's self-attention, we compare with Big Bird\cite{zaheer2020big} and Reformer\cite{kitaev2020reformer}. We keep other experimental settings like model's embedding vector size and hyperparameters same and only modify the attention mechanism for fair comparisons.

%\betterparagraph{Implementation Details}

% We have implemented \proposed{}'s optimized sparse attention using CUDA and compiled it into a shared object. By leveraging CUDA's parallel processing capabilities on GPUs, we can perform accelerated computations for sparse attention operations. This shared object has been integrated with PyTorch.
% To incorporate our CUDA sparse attention into the training process of a PyTorch model, we have created a custom autograd function in PyTorch. To ensure effective utilization of the PyTorch custom autograd function, we have implemented the forward and backward CUDA functions specifically for our sparse attention module.

% \gm{We implemented a 2-layer and 2-head encoder-only Transformer model} for our evaluations, where embedding dimension ($D$) is set to 64 for all experiments.

In our \proposed{} implementation, we used an embedding dimension ($D$) of size 64, and the batch size was chosen based on the available memory size, resulting in batch sizes of 256 for image classification, 128 for ListOps and 32 for document retrieval.
%and the feed-forward network hidden states dimension size is set to 128.
%Finally, for classification purposes, we utilized mean pooling for the output of the model. 
% The batch size was selected based on the memory requirements, resulting in batch sizes of 256 for image classification, 128 for ListOps and 32 for document retrieval.
%We used a one-cycle learning rate scheduler with a learning rate of 5e-5 for text classification, and a learning rate of 2e-4 for image classification, ListOps and document retrieval.
%For Text Classification and Document Retrieval tasks, each model was trained for 30k steps, while the other tasks were trained for 50k steps. 
%During training, the best checkpoint with the highest accuracy on the development set was saved for evaluation. 
%The number of dense steps for \proposed{}, which refers to the number of steps before generating the sparsity pattern in the attention score matrix, was determined by observing the convergence of the model during the training process. Specifically, we used 3,000 steps, 2,500 steps, 1,000 steps, and 2,500 steps for image classification, ListOps, text classification and document retrieval, respectively. 
The size of convolution filter in \proposed{} is set to $(31 \times 31)$ for all experiments. And for the threshold used in the flood filling method, we set $\alpha$ to 96 for image classification, 98 for ListOps, and 99 for document retrieval tasks.
While performing average pooling and upsampling, the block size for our model was determined based on the maximum sequence length of the respective dataset. For image classification task, we used a block size of 32 and for ListOps and document retrieval tasks, a block size of 64 is used. 
Note that all the experiment results presented in this section are averaged over 3 different executions.

\subsection{Performance Evaluation}

% \begin{figure}[!ht]
% \centering
%   \includegraphics[width=0.49\textwidth]{figures/convergegraph.pdf}
%   \caption{Comparison of validation accuracy over training time on four tasks.X-axis: training time, Y-axis: classification accuracy}
%   \label{fig:converge}
% \end{figure}
% \index{figure}

\begin{table}[ht]
\centering
\caption{Comparison of accuracy (\%) on three LRA tasks}
\scalebox{0.9}{
\begin{tabular}{|c|c|c|c|c|}
\hline
       Model     & \begin{tabular}[c]{@{}c@{}}Image\\ Classification\end{tabular}           & ListOps                   & \begin{tabular}[c]{@{}c@{}}Document\\ Retrieval\end{tabular}       \\ \hline
Original    & 39.022          & 38.615                 & 77.796          \\ \hline
BigBird     & 38.927          & 38.722                 & 78.437          \\ \hline
Reformer    & 40.488          & 36.535          & 75.657          \\ \hline
\proposed{}-C & 39.173          & 38.889                   & 76.015          \\ \hline
\proposed{}-F   & 40.926          & 38.889                & 79.120          \\ \hline
\proposed{}-CF      & \textbf{41.313} & \textbf{38.958}         & \textbf{79.238} \\ \hline
\end{tabular}
}
\label{tab:lraacc}
\end{table}

% \begin{table}[ht]
% \centering
% \scalebox{0.92}{
% \begin{tabular}{|c|c|c|c|c|}
% \hline
%        Model     & \begin{tabular}[c]{@{}c@{}}Image\\ Classification\end{tabular}           & ListOps         & \begin{tabular}[c]{@{}c@{}}Text\\ Classification\end{tabular}            & \begin{tabular}[c]{@{}c@{}}Document\\ Retrieval\end{tabular}       \\ \hline
% Original    & 39.022          & 38.615          & 63.642          & 77.796          \\ \hline
% BigBird     & 38.927          & 38.722          & 63.567          & 78.437          \\ \hline
% Reformer    & 40.488          & 36.535          & \textbf{63.872} & 75.657          \\ \hline
% \proposed{}-Conv & 39.173          & 38.802          & 61.873          & 76.015          \\ \hline
% \proposed{}-FF   & 40.926          & 38.889          & 63.363          & 79.120          \\ \hline
% \proposed{}      & \textbf{41.313} & \textbf{38.958} & 63.175          & \textbf{79.238} \\ \hline
% \end{tabular}
% }
% \caption{Accuracy (\%) of Long Range Arena tasks}
% \label{tab:lraacc}
% \end{table}
\betterparagraph{Convergence}
%Figure \ref{fig:converge} shows the validation accuracy over training time for four different tasks. 
%\proposed{} consistently outperformed existing state-of-the-art models on Image Classification and ListOps tasks. 
% ? produced a better convergence rate than other models. 
% ? converged faster
%BigBird model suffered from a lower convergence on both Image Classification and ListOps tasks. 
Table \ref{tab:lraacc} shows the accuracy of six models on three different tasks. 
Our \proposed{}-CF consistently achieved the highest accuracy in all tasks, surpassing the highest accuracy obtained from other compared models by +0.825\%, +0.236\%, and +0.801\% for the three evaluation tasks.
% with +0.825\%, +0.236\%, +0.801\% against the highest accuracy of other compared models. 
% Among \proposed{} variants, 
It is interesting to see that among the \proposed{} variants, incorporating both the convolution filter and flood-filling scheme led to higher accuracy for all tasks.
% Combining both the convolution filter and flood filling method can result in a synergistic effect. 
This indicates that the convolution filter and flood-filling method synergize with each other.
More specifically, in the sparsity pattern generation phase, the convolution filter plays a role in increasing the values of distinct shapes appearing in $A^s$, thereby enhancing the ability of the flood filling method to identify critical elements more effectively.
Additionally, we observed that the flood filling method shows more significant effect on accuracy compared to the convolution filter.
This result demonstrates that, in sparsity pattern generation, it is more important to consider the connectivity between elements than to simply select elements based on their values.
%As the flood filling method take account for the identifying the critical elements in A^s considering the connectivity, selecting top 몇개 보다 generating sparsity pattern에 impact 있다.
BigBird showed relatively lower accuracy for image classification and relatively higher accuracy for document retrieval.
% This is because the BigBird model was specifically  developed for addressing long sequences of text dataset.
This is because the BigBird model was specifically developed to address long sequences of text datasets.

\begin{figure}[!ht]
\centering
\includegraphics[width=0.49\textwidth]{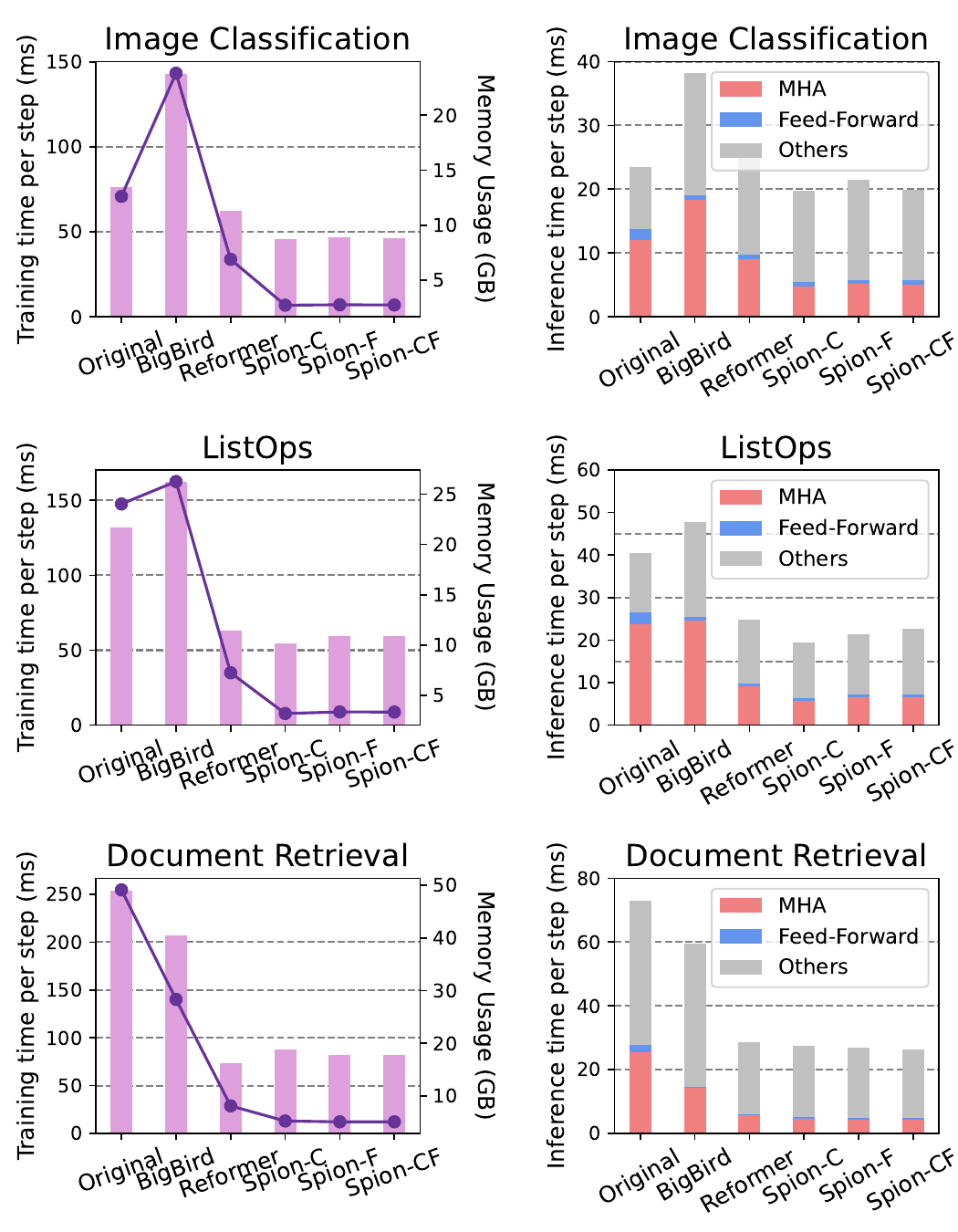}
% \caption{Memory usage and training time results on three tasks. We report the average memory usage (GB) and training time (ms) for one training step. In order to ensure a fair comparison, we divided the total training time and total memory usage by the total number of steps. Bar graphs: training time, line graphs: memory usage.}
\caption{Comparison of elapsed time and memory footprint per step for training (left column), and inference time per step (right column) for three tasks.
% We report the average memory usage (GB) and training time (ms) for one training step. 
% In order to ensure a fair comparison, we divided the total training time and total memory usage by the total number of steps. Bar graphs: training time, line graphs: memory usage.
}

\label{fig:traingraph}
\end{figure}
\index{figure}

%%%%%%%%%%%%%%%%%%%%%%%%%%%%%%%%%%%%
%\input{tables/inftime}
%\input{tables/mhatime}

%or 

% \begin{figure}[!ht]
% \centering
% \includegraphics[width=0.49\textwidth]{figures/MHAbreakdown.pdf}
% \caption{Breakdown of 
% elapsed time (ms) per step during model inference. 
% \gm{Elapsed time per step for inference}
% Others include input embedding operation, position embedding operation and sequence classifier for the last output layer.}
% \label{fig:mhabreakdown}
% \end{figure}
% \index{figure}

%%%%%%%%%%%%%%%%%%%%%%%%%%%%%%%%%%%%%%

\betterparagraph{Speedup}
Figure \ref{fig:traingraph} shows the time and memory space required for training, as well as the time required for processing inference on three tasks.
Compared to the original Transformer model, our \proposed{}-CF achieved $1.66\times$, $2.21\times$ and $3.08\times$ speedup per training step on each task.
\proposed{} achieved significant speedup, particularly for tasks involving longer sequences, such as document retrieval. 
As the input sequence length increases, the number of operations in MHA increases exponentially. Consequently, in models that handle longer sequences, the proportion of operations accounted for MHA may be larger compared to models dealing with shorter sequences. 
Therefore, in longer sequences, reducing the count of operations involved in the MHA leads to higher performance.
%achieving a significant speedup can be accomplished by reducing the number of operations of MHA that contribute the most to the overall computation in the model.
%Noting that \proposed{} has a phase of utilizing the dense attention mechanism at the beginning of the training process, it still achieved the shortest training time in \bk{most?} tasks.
% ? is the fastest of all competing implementations.
% Note that \proposed{} variants, including \proposed{}-C and \proposed{}-F, are not closely related to the speedup or memory footprint, all three \proposed{} variants show similar training times for all tasks. 
The figure on the right column in Figure \ref{fig:traingraph} shows a breakdown of inference time of compared models.
Compared to the original Transformer, for document retrieval task, our \proposed{}-CF achieved a speedup of $5.54\times$ for the executing MHA operation while achieving a speedup of $2.78\times$ in total elapsed time for inference.
% With our \proposed{}, we significantly reduce the number of computations in MHA, which takes the largest proportion of overall operations. And this leads to a significant decrease in the total number of operations.

% $2.84\times$, $4.04\times$, and $5.54\times$ 
% , while achieving a speedup of $1.18\times$, $1.78\times$, and $2.78\times$ in total elapsed time for each task.
%In the Original model, the MHA operation takes the largest proportion of overall operations, while in the proposed model, MHA no longer occupies the greatest proportion, leading to a significant decrease in the total number of operations.
% With our \proposed{}, we significantly reduce the number of computations in MHA, which takes the largest proportion of overall operations. And this leads to a significant decrease in the total number of operations.
% Reducing the computational load of the MHA operation, which accounts for the largest portion of overall model operations, leads to a significant decrease in the total number of operations. 
%Since the time taken by Multi-Head Attention (MHA) has decreased, it is no longer a bottleneck for the model.
%\proposed{} achieved a significant speedup in model inference for tasks involving longer sequence lengths, such as document retrieval.

\begin{figure}[!ht]
\centering
\includegraphics[width=0.4\textwidth]{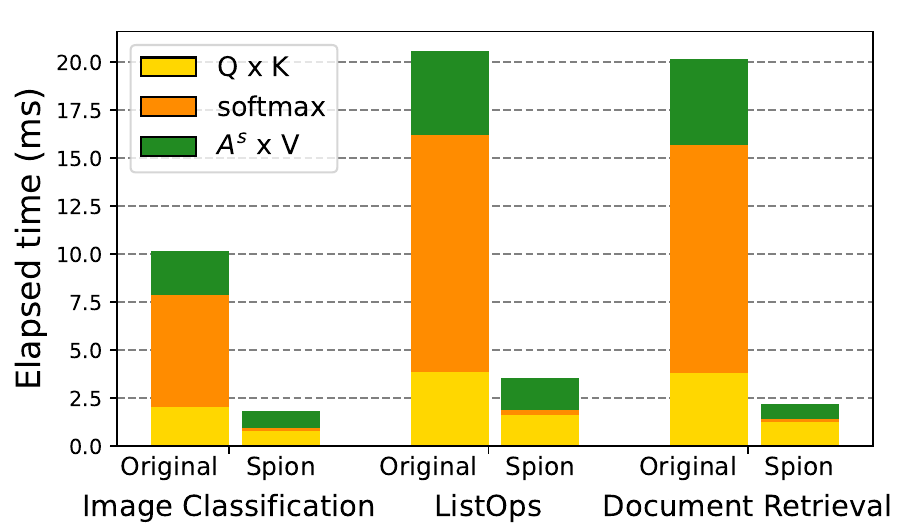}
\caption{
Breakdown of elapsed time (ms) for running MHA operations on three tasks
% Elapsed Time (ms) Breakdown of Each Operations of the MHA
}
\label{fig:attnbreakdown}
\end{figure}
\index{figure}
%each operations in Attention (vs. Original) with breakdown 

To evaluate the reduction in elapsed time for each operation in the dense MHA and sparse MHA, we compared the elapsed time of each operation in our sparse MHA with that of the original dense MHA.
Figure \ref{fig:attnbreakdown} shows a breakdown of the elapsed time for running MHA operations in the original Transformer and \proposed{} on three different tasks.
For the image classification task, the replacement of the GEMM operation of $Q$ and $K$ with the SDDMM operation resulted in a speedup of $2.55\times$.
% As shown in Figure \ref{fig:attnbreakdown}, in the image classification task, the replacement of the GEMM operation of $Q$ and $K$ with the SDDMM operation resulted in a speedup of $2.55\times$. 
The softmax operation achieved a significant speedup of $42.40\times$. 
Additionally, the SpMM operation, which replaced the GEMM operation for $A^s$ and $V$, outperformed with a speedup of $2.54\times$.
This result demonstrates that the softmax function implemented in the original Transformer model is a primary bottleneck. 
% However, as our \proposed{} model performs the optimized softmax by considering the sparsity, the execution time for softmax operation is greatly reduced.
However, since our \proposed{} model performs optimized softmax by considering sparsity, the execution time for the softmax operation is significantly reduced.
%Replacing the softmax function with sparse softmax achieved the most significant speedup among the three operations.
Not only for the image classification task, but also for all other tasks, our \proposed{} achieved speedup in every operation associated with the MHA.
%After replacing softmax function with the sparse softmax function, the softmax function has become no longer a bottleneck.
% Original attention mechanism  is dominated by ? operation as shown in ?

%Table ? shows the elapsed time breakdown for each step in attention mechanism. 

\begin{figure}[!ht]
\centering
\includegraphics[width=0.4\textwidth]{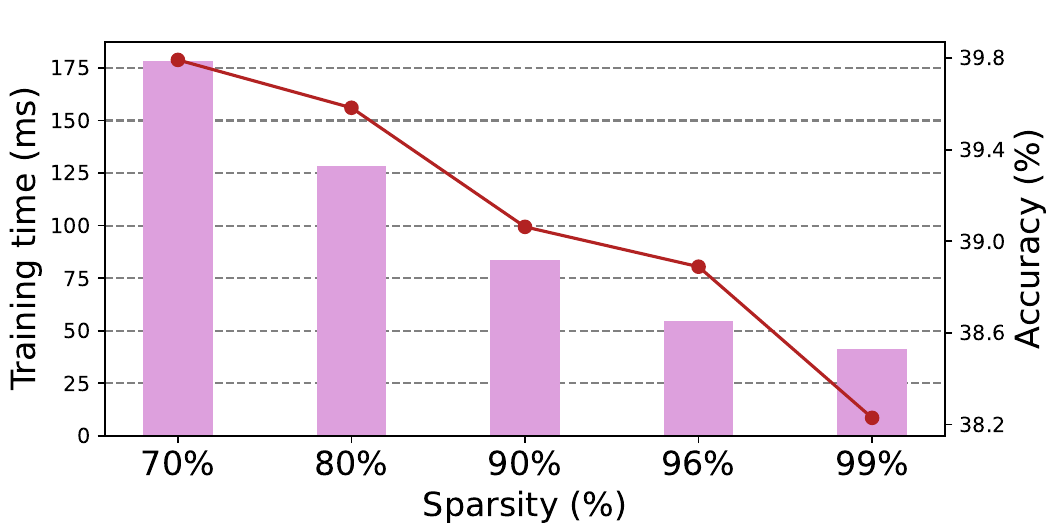}
\caption{
Comparison of training time and accuracy over different sparsity ratio
% Right: Comparison of training time per step and accuracy over the sizes of block.
% results by block size in the image classification task.
% classification accuracy and training time results by density in the ListOps task
% Left: Comparison of classification accuracy and training time results by density in the ListOps task. Right: Comparison of classification accuracy and training time results by block size in the image classification task. X-axis: density, left Y-axis: training time (ms), right Y-axis: classification accuracy (\%). Bar graphs: training time, line graphs: memory usage.
}
\label{fig:sparsitygraph}
\end{figure}
\index{figure}

% \betterparagraph{Comparing Speedup and Accuracy across Sparsity Ratio}
\betterparagraph{Sparsity Ratio Comparison}
Figure \ref{fig:sparsitygraph} shows the training time and classification accuracy at each sparsity ratio on the ListOps task.
Since it is not possible to adjust the sparsity in the flood-filling scheme, we adopted the \proposed{}-C model to evaluate the impact of sparsity ratio.
As expected, a higher sparsity ratio leads to shorter training times, but it results in lower accuracy. 
Thus, it is crucial to find the optimal sparsity ratio that balances between achieving high accuracy and reducing training time. 
In the ListOps task, a sparsity ratio of 96\% provides significantly reduced training time while maintaining a comparable accuracy. 
Compared to the results of a sparsity ratio of 70\%, achieving a 96\% sparsity ratio resulted in a speedup of $3.26\times$, while the accuracy only dropped by 0.903\%, which is less than 1\%.

%On right of Figure \ref{fig:densitygraph} shows the comparison of classification accuracy and training time of Multi-Head Attention (MHA) between different block size.
%We used block sizes of 16, 32, 64, and 128 for the image classification task to analyze the impact of block size on classification accuracy and training time.
%We evaluated for image classification tasks with block size of 16, 32, 64 and 128.
%Increasing the block size enhances data locality, leading to reduced data movement and decreased training time. However, it is important to note that as the block size increases beyond a certain point, the relationships between neighboring elements may become less evident, leading to reduced accuracy.

\betterparagraph{Memory Footprint}
As shown in Figure \ref{fig:traingraph}, our \proposed{}-CF achieved a memory footprint reduction of $4.62\times$, $7.23\times$, and $9.64\times$ compared to the original Transformer model. 
And our \proposed{} variant models resulted the least memory footprint in all tasks. 
Compared to Reformer, our \proposed{} demonstrates comparable training time but outperforms it in terms of memory footprint.
This result shows that our \proposed{} is able to efficiently perform sparse MHA using sparsity patterns without requiring additional memory.
% We apparently showed that \proposed{} can efficiently perform sparse MHA using sparsity patterns without requiring additional memory.
In addition, \proposed{}-CF achieved the highest accuracy across all tasks, demonstrating that the achieved memory savings do not come at the expense of accuracy.

\section{Conclusion}
\label{sec:conclusion}
Due to the compute-intensive nature of the Transformer, that must perform a large number of MHA operations, we present a novel sparsification scheme that leverages convolution filters and the flood-filling algorithm to dynamically reduce the number of operations in the MHA.
Our sparsification technique is capable of identifying various aspects of the sparsity pattern in the MHA operation and is applicable to many other Transformer models, not limited to the encoder-only Transformer.
Furthermore, we developed a parallel sparse MHA implementation to achieve high performance.
Experimental results on long sequence datasets demonstrate that our parallel implementation on GPUs achieves a significant performance improvement over state-of-the-art sparsity-aware Transformer models, while maintaining comparable or better accuracy.

% \rev{Replace or add more references recently published in top-tier conferences, so many arXiv papers in the current list}
\bibliographystyle{ACM-Reference-Format}
\bibliography{references}

% \appendix

\end{document}